\pdfoutput=1

\documentclass[11pt]{article}

\usepackage[final]{acl}

\usepackage{multirow}
\usepackage{xcolor}
\usepackage{makecell}
\usepackage{url}

\usepackage{times}
\usepackage{latexsym}
\usepackage{amsmath}
\usepackage{fontawesome5}
\usepackage{subcaption}

\usepackage[T1]{fontenc}

\usepackage[utf8]{inputenc}

\usepackage{bbm}
\usepackage{microtype}
\begingroup
\lccode`\~=`\|
\lowercase{\endgroup
\def\tup#1{{\def~{\;\middle\vert\;}\mathcode`\|="8000\left\langle#1\right\rangle}}}
\usepackage{tabularx}
\usepackage{array} 

\usepackage{inconsolata}

\usepackage{graphicx}

\title{Contextual Metric Meta-Evaluation by Measuring Local Metric Accuracy}

\author{Athiya Deviyani \\
 Carnegie Mellon University \\
 \texttt{adeviyan@cs.cmu.edu} \\\And
 Fernando Diaz \\
 Carnegie Mellon University \\
 \texttt{diazf@acm.org} \\}

\begin{document}
\maketitle
\begin{abstract}
Meta-evaluation of automatic evaluation metrics---assessing evaluation metrics themselves---is crucial for accurately benchmarking natural language processing systems and has implications for scientific inquiry, production model development, and policy enforcement. While existing approaches to metric meta-evaluation focus on general statements about the absolute and relative quality of metrics across arbitrary system outputs, in practice, metrics are applied in highly contextual settings, often measuring the performance for a highly constrained set of system outputs. For example, we may only be interested in evaluating a specific model or class of models. We introduce a method for contextual metric meta-evaluation by comparing the \textit{local metric accuracy} of evaluation metrics. Across translation, speech recognition, and ranking tasks, we demonstrate that the local metric accuracies vary both in absolute value and relative effectiveness as we shift across evaluation contexts. This observed variation highlights the importance of adopting context-specific metric evaluations over global ones.

\end{abstract}

\section{Introduction}
\label{sec:intro}

Meta-evaluation of automatic evaluation metrics---assessing evaluation metrics themselves---is crucial for accurately benchmarking natural language processing systems \cite{zhou:deconstructing:naacl}. Because metrics are central to scientific inquiry, the development of production models, and policy enforcement \cite{kocmi:ship:wmt}, there is a constant need for new approaches to rigorously evaluate these metrics to ensure they remain reliable and contextually appropriate, supporting their effective use across diverse NLP tasks and evolving systems \cite{novikova:need:acl}.

Although current methods for metric meta-evaluation commonly take a \textit{global} perspective, reporting the performance of a metric across arbitrary system outputs, coming from any system \cite{stanojevic:wmt2015:wmt, przybocki:nist:mt}, in practice, evaluation is highly contextual, measuring the performance for a highly constrained set of system outputs. The evaluation \textit{context} refers to any meaningful subset of evaluation data, with the constrained set of system outputs being one specific type of context (e.g., outputs from particular models or specific levels of output quality). For example, we may only be interested in evaluating a specific model or class of models. 
From a model development perspective, a metric that is sensitive to model outputs coming from partially trained models at the beginning of the development cycle (when the outputs are far from the target distribution or close to random) may struggle to differentiate between outputs from fully trained or more effective models. For example, \citet{fomicheva:mteval:cl} show that the performance of a metric changes as the translation quality changes. Thus, using the same metric across various contexts will have inconsistent reliability and can result in inaccurate model selection.

\begin{table}
 \centering
 \begin{tabular}{c|ccc|c} 
   \multicolumn{1}{c}{}& \multicolumn{3}{c}{context}\\ 
   metric& \textbf{X}& \textbf{Y}& \textbf{Z} & \textbf{global}\\ \hline 
   A& 0.9& 0.9& 0.3& 0.7\\ 
   B& 0.7& 0.7& 0.7& 0.7\\ 
   C& 0.3& 0.3& 0.9& 0.5
 \end{tabular}
 \caption{Contextual metric meta-evaluation. The values in the table are the metric accuracies, which represents how often a metric correctly estimates the true ordering of a pair of systems. When comparing metrics A, B, and C, traditional meta-evaluation focuses on global accuracy across arbitrary inputs. Local metric accuracy can vary by evaluation contexts X, Y, and Z.}
 \label{tab:toy_example}
\end{table}

To illustrate the difference between global and contextual metric meta-evaluation, we constructed a toy meta-evaluation for three metrics across three contexts (Table \ref{tab:toy_example}). The values in the table represent the accuracy of three metrics (A, B, and C) under three different contexts (X, Y, and Z). By looking at the average, we might think that A and B are equally accurate. However, when inspecting accuracy within individual contexts, we can see that selecting the most appropriate metric is far less straightforward than simply choosing based on the highest global accuracy, as it fails to capture the metric's variability and effectiveness in specific contexts. For example, if we want a metric that best generalizes across different contexts, we want to choose B over A for its robustness, even though their global accuracies are equal. However, if we want to specifically measure outputs in context Z, then we would want to pick C as it is especially sensitive to system outputs in that context despite it having the lowest global accuracy.

Although prior research has focused mainly on global metric evaluations, we explicitly analyze how metric accuracy varies across different evaluation contexts by measuring their \textit{local metric accuracies}. By evaluating metrics across three different machine learning tasks---machine translation, automatic speech recognition, and ranking---we show that metric accuracy, which measures the ability of a metric to accurately assign the true preference between a pair of system decisions, changes as evaluation context changes. Specifically, we show that the metric accuracy changes both in absolute value and relative ordering across the different contexts. In contrast with existing work on metric meta-evaluation, which relies heavily on costly and time-consuming explicit human feedback \cite{fabbri:summeval:acl, liu:dialeval:emnlp}, our method uses output perturbations \cite{sai:checklists:emnlp, he:blindperturb:acl} to obtain the true ordering between a pair of system outputs without the need for human supervision. Overall, we show that measuring local metric accuracies is a straightforward methodology to provide a more contextual understanding of evaluation metrics which complements existing global metric meta-evaluation methods. 

\section{Related work}
\label{sec:related_work}

Our work is situated in the broader literature on meta-evaluation. For example, the Conference on Machine Translation (WMT) has focused on evaluating the utility of metrics in machine translation since 2008, where participants submit automatic metrics for validation against human feedback \cite{ccb:wmt:acl}. \citet{xiao:evaluating:emnlp} propose a meta-evaluation framework rooted in measurement theory for NLG metrics, highlighting issues in human evaluation that include a lack of validation, standardization, and consistency. 

Our use of output perturbations is inspired by prior work in testing metric robustness. \citet{chen:menli:tacl} proposed a preference-based adversarial attack framework using targeted perturbations to evaluate the robustness of NLI-based and BERT-based metrics, finding that NLI-based metrics are more robust in summarization but not in machine translation. \citet{sai:checklists:emnlp} extends perturbation-based robustness testing by creating templates that target specific criteria such as jumbled word order to test fluency. \citet{he:blindperturb:acl} utilized perturbations to design synthetic stress tests to evaluate the robustness of various text generation metrics. Furthermore, \citet{valcarce:topn:acmrecsys} evaluated the robustness of the ranking metrics against incompleteness by purposefully removing system outputs to introduce sparsity. While these works have demonstrated the utility of perturbations for evaluating metric robustness, our paper introduces a novel method for contextual metric meta-evaluation that leverages perturbations as a methodological tool to obtain reliable preferences between system outputs without requiring costly and time-intensive human annotations.

Although metric meta-evaluation is often done on a global level, previous work indicates that the reliability of a metric changes from the system-level to the decision-level \cite{reiterbelz:sysok:acl, stent:sentnotok:springer}. Though some research has investigated metric performance for different contexts based on output sources (i.e., models) or output qualities \cite{mathur:wmt2020:wmt, novikova:need:acl}, our work addresses the lack of a systematic review of contextual meta-evaluation and how to conduct it. Recent studies have further motivated the need for contextual metric meta-evaluation by highlighting the lack of robustness of metrics across varying conditions. For instance, \citet{falcao:cometlowmt:lrec} showed that COMET's performance varies differently as we move from high-resource to low-resource languages. \citet{zouhar:finetuned:acl} also showed that the performance of fine-tuned metrics substantially drops in unseen domains. Additionally, \citet{zouhar:quality:lrec} showed that the number of higher-quality references significantly impacts the reliability of a metric. These findings highlight the importance of context-specific evaluations, as metrics that perform well globally may fail in specific settings, such as low-resource languages or unseen domains. Challenge sets like the one introduced by \citet{amrhein:aces:wmt} further emphasize the need for context-aware evaluation metrics, as their findings show that no single metric consistently won across all error categories. These works highlight the critical gap our contextual meta-evaluation method aims to address, providing a systematic way to better understand metric performance across diverse evaluation scenarios.

\section{Local accuracy}
\label{sec:problem_definition}

To formalize local metric accuracy, we introduce the following notation. Let $\mathcal{X}$ be the set of all possible system inputs (e.g., for MT, all possible strings from the source language) and $\mathcal{Y}$ the set of all possible system outputs (e.g., for MT, all possible strings from the target language). We define $X\subset\mathcal{X}$ to be the subset of system inputs observed in a specific context (e.g., for MT, a sample of source sentences from a specific university). Similarly, $\mathcal{Y}_x\subset\mathcal{Y}$ is the subset of system decisions for $x\in X$ observed for $X$ in a specific context (e.g., for MT, a set of translations generated by a set of candidate systems). In addition, we have access to several perturbation functions that, with high probability, degrade the utility of a decision $y$ (e.g., dropping a random word from a translated input). Let $Q_x$ be the set of pairs decisions conditioned on an input $x$ and their corresponding degraded version: $Q_x=\{\tup{y,y'}\}_{y\in \mathcal{Y}_x}$.

An evaluation metric $\mu: \mathcal{X}\times\mathcal{Y}\rightarrow \Re$ generates a scalar number reflecting the performance according to some system property that we want to measure (e.g., correctness of translation). Each metric is an approximation of $\mu^*$, the unobserved ideal evaluation metric (i.e., the true utility of an output). Given two pairs of system outputs, $\mu^*$ will always be able to determine the true ordering of the two outputs, including instances where it has no preference. In cases where we intentionally perturb $y$ to obtain $y'$, we have that $\mu^*(x,y) > \mu^*(x,y')$. We can assume that the system outputs that we are evaluating are at least better than random such that we can assume that $\mu^*(x,y) > \mu^*(x,y')$ with high probability. Under the assumption that $\mu$ approximates $\mu^*$, we want to compute how often $\mu(x,y) > \mu(x,y')$. As suggested by \citet{kocmi:ship:wmt}, we focus on the ability of $\mu$ to reproduce the ordering of decisions rather than the magnitude of the difference between $\mu(x,y)$ and $\mu(x,y')$. From this, we define the pointwise local metric accuracy, conditioned on an input $x$, to be:

\small
\begin{align}
\label{eq:contextacc}
\textsc{Acc}_{\mu}(Q_x) &= \frac{1}{|Q_x|} \sum\limits_{\tup{y,y'} \in Q_x} \mathbbm{1}\left[\mu(x,y) > \mu(x,y')\right]
\end{align}
\normalsize

This measures the ability of a metric to reproduce the true ordering of perturbations for a specific input $x$. We define the local metric accuracy across all inputs $X$ by averaging over $|X|$, as follows:
\begin{align}
\label{eq:localacc}
\textsc{Acc}_{\mu}(Q) &= \frac{1}{|X|} \sum_{x \in X} \textsc{Acc}_{\mu}(Q_x)
\end{align}
where $Q=\cup_{x\in X} Q_x$. This measures the local metric accuracy across a sample of system inputs, as we may have in a standard evaluation set. The main difference with global metric accuracy is that the accuracy is only computed over all pairs of $\tup{y,y'}$ belonging to a specific context (and not over all the observed outputs).

We are interested in testing two hypotheses with respect to local metric accuracy. 

\begin{enumerate}
 \item[H1:] The \textit{absolute} local metric accuracy, $\textsc{Acc}_{\mu}(Q)$, of a metric $\mu$ changes as the context changes. 
\end{enumerate}
Evidence supporting this hypothesis suggests that existing evaluation methods focusing on global metric accuracy obscure how metric accuracy varies across different contexts.
\begin{enumerate}

 \item[H2:] The \textit{relative} local metric accuracy of a metric $\mu$ changes as the context changes.
\end{enumerate}
In other words, the total ordering of all metrics by local metric accuracy within a context changes as the context changes. 
Evidence supporting this hypothesis suggests that choosing an appropriate metric to benchmark compare system outputs largely depends on the context.

\section{Methods and Materials}
\label{sec:experimental_setup}

\subsection{Tasks, dataset, and metrics}
\label{sec:dataset_tasks}
We performed our evaluation on three different tasks: machine translation (MT), automatic speech recognition (ASR), and ranking. We detail the dataset, metrics, and their corresponding implementations in Appendix \ref{sec:materials_details}.

We can divide the outputs for each task into different subsets, such as the \textsc{System} that produced it. We report results on the subset with the highest number of contexts for each task. The abundance of contexts allowed us to identify trends in metric behavior across a broader range of items and helped us identify supporting evidence for or against our hypotheses. We have included the results for each of the available subsets for each task in Appendix \ref{sec:other_categories}.

\subsection{Perturbation techniques}
\label{sec:perturbation_techniques}
\begingroup
\setlength{\tabcolsep}{6pt} 
\renewcommand{\arraystretch}{1.5} 
\begin{table*}[t]
{\small
\centering
\begin{tabular}{lp{0.9in}p{4.5in}}
\textbf{Task} & \textbf{Perturbation} & \textbf{Description} \\ \hline
\multirow{3}{*}{MT} 
 & Removal & Drop a random word from the string, e.g. ``John said he did not like the cake baked.'' \\ 
 & Swapping & Switch the position of a random pair of words from the string, e.g.``John \colorbox{pink}{he} \colorbox{pink}{said} did not like the cake she baked.'' \\
 & LLM & Prompt an LLM to perturb the string, e.g. ``John \colorbox{pink}{mentioned} he \colorbox{pink}{didn’t} \colorbox{pink}{enjoy} the cake she \colorbox{pink}{made}.''\\ \hline
\multirow{2}{*}{ASR} 
 & Phonetic sub. & Replace a random letter (or a letter group) with phonetically similar ones, e.g. ``John \colorbox{pink}{sed} he did not like the cake she \colorbox{pink}{bakked}.'' \\
  & LLM & Prompt an LLM to perturb the string, e.g. ``John \colorbox{pink}{sed} he did not like cake she baked.'' \\ \hline
\multirow{1}{*}{Ranking} 
 & Swapping & Switch the position of an item with another item within the ranked list, e.g. Original ranking: [Cake, Pie, Cookie] Perturbed: [Cookie, Pie, Cake]\\ \hline
\end{tabular}}
\caption{A subset of perturbations for MT, ASR, and Ranking tasks. The original unperturbed sentence for the MT and ASR examples is ``John said he did not like the cake she baked.'' We show the full table of perturbations in Appendix \ref{sec:perturbation_appendix} and outline how each perturbation method is implemented in Appendix \ref{sec:perturbation_implementation}.
}
\label{table:mini_perturbations}
\end{table*}
\endgroup

To test our hypotheses, we apply perturbation functions that degrade the utility of a system output $y$ to obtain its corresponding degraded version $y'$, simulating the output of a degraded system. This methodology is widely used to automatically generate pairs of outputs with their corresponding true ordering for meta-evaluation, such as in work done by \citet{he:blindperturb:acl} and \citet{sai:checklists:emnlp}. By applying perturbations, we know that the quality of $y'$ under a specific task is worse than $y$ with high probability. We highlight a subset of the perturbations applied to the system outputs from their corresponding tasks in Table \ref{table:perturbations}. The full list of perturbations can be found in Appendix \ref{sec:perturbation_appendix}. We selected this set of perturbations because it represents a large enough breadth yet still mimics realistic system errors in their corresponding tasks. It is important to note that not all perturbations apply to all outputs. For example, some outputs do not contain named entities, so we cannot apply the perturbation method involving a named entity substitution.

Certain perturbations may confound the results, i.e., they may inherently favor (or disfavor) specific metrics due to their design. For example, if we apply the swapping perturbation function, this might disproportionately penalize \textsc{Rouge-1} as it is designed only to measure lexical overlap and is agnostic to position; therefore, it will consistently achieve an accuracy of 0 across all contexts. However, we likely do not observe this behavior from \textsc{Rouge-1} under different perturbations, such as removal or insertion. Thus, if a perturbation confounds a metric, the local metric accuracies will exhibit high variance when we change the perturbation method.

Finding a single perturbation method that does not confound all the metrics we evaluate is challenging. To mitigate the impact of potential confounders, we generate four different $y'$s for each $y$ in machine translation and automatic speech recognition, corresponding to the result of applying four randomly selected perturbations to $y$. Let $\{f_1, f_2, f_3, f_4\}$ be the randomly selected perturbations corresponding to any of the perturbation methods for a specific task. Now, instead of having a single pair $\tup{y,y'}$ in $Q_x$, we will have four pairs for each $y$, namely $\tup{y,f_1(y)}$, $\tup{y,f_2(y)}$, $\tup{y,f_3(y)}$, and $\tup{y,f_4(y)}$. Since the different perturbation methods are independent, increasing the number of samples for $y'$ by applying different perturbations reduces the variance in local metric accuracies across various combinations of perturbations. This, in turn, minimizes the effect of confounders. We illustrate this further in our experiments in Appendix \ref{sec:more_perturbations_good}.

Given the inherent structure of ranked lists, our approach involves applying a single perturbation method for the system outputs in the ranking task, as there are limited meaningful ways to modify a ranked list to simulate realistic system errors. To address this limitation, we swap multiple items within the list when applying the swapping perturbation to ranked outputs. In contrast, for machine translation and automatic speech recognition, we apply four different perturbation methods to a single output $y$ to get four distinct pairs of $\tup{y,y'}$, but we apply each method only once (e.g., when applying the swapping perturbation, we only swap a pair of words instead of swapping multiple pairs).

\subsection{Hypothesis Testing}

We plotted the metric accuracies $\textsc{Acc}_{\mu}(Q)$ for each task across different contexts within the selected context subset $Q$ as a line graph, such that we can visualize how the metric's capability of differentiating between $y$ and $y'$ changes as the context changes by observing the slopes and overlaps between the lines. To test H1 and to further investigate the association between the context $Q$ and the metric accuracy $\textsc{Acc}_{\mu}(Q)$, we used the $\chi^2$ test of independence of variables \cite{pearson:chi2:philmag} in a contingency table \cite{pearson:contingencytable:dulau}. We compared the resulting $p$-values to the significance level of $\alpha = 0.05$ to understand whether the changes in metric accuracy $\textsc{Acc}_{\mu}(Q)$ across the different contexts $Q$ are statistically significant.

 To test H2, we computed Kendall's-$\tau_{AP}$ \cite{kendall:kendalltau:biometrika} between two rankings of metrics according to local metric accuracy under two contexts. The Kendall's-$\tau_{AP}$ values help us quantify how the metrics' total ordering changes as the context changes. To emphasize the metric selection task, we adopt a widely-used variant of Kendall's-$\tau_{AP}$ known as $\tau_{AP}$ proposed by \citet{yilmaz:tauap:sigir}, such that changes higher on the ranked list are weighted more than the ones that are much lower in the list. We will also compare the rankings of metrics according to local metric accuracy within one context and the rankings of metrics according to their global metric accuracy.

\section{Results}
\label{sec:results}

\begin{figure*}
    \centering
    \begin{subfigure}[t]{0.49\textwidth}
    \centering    
    \includegraphics[width=\textwidth]{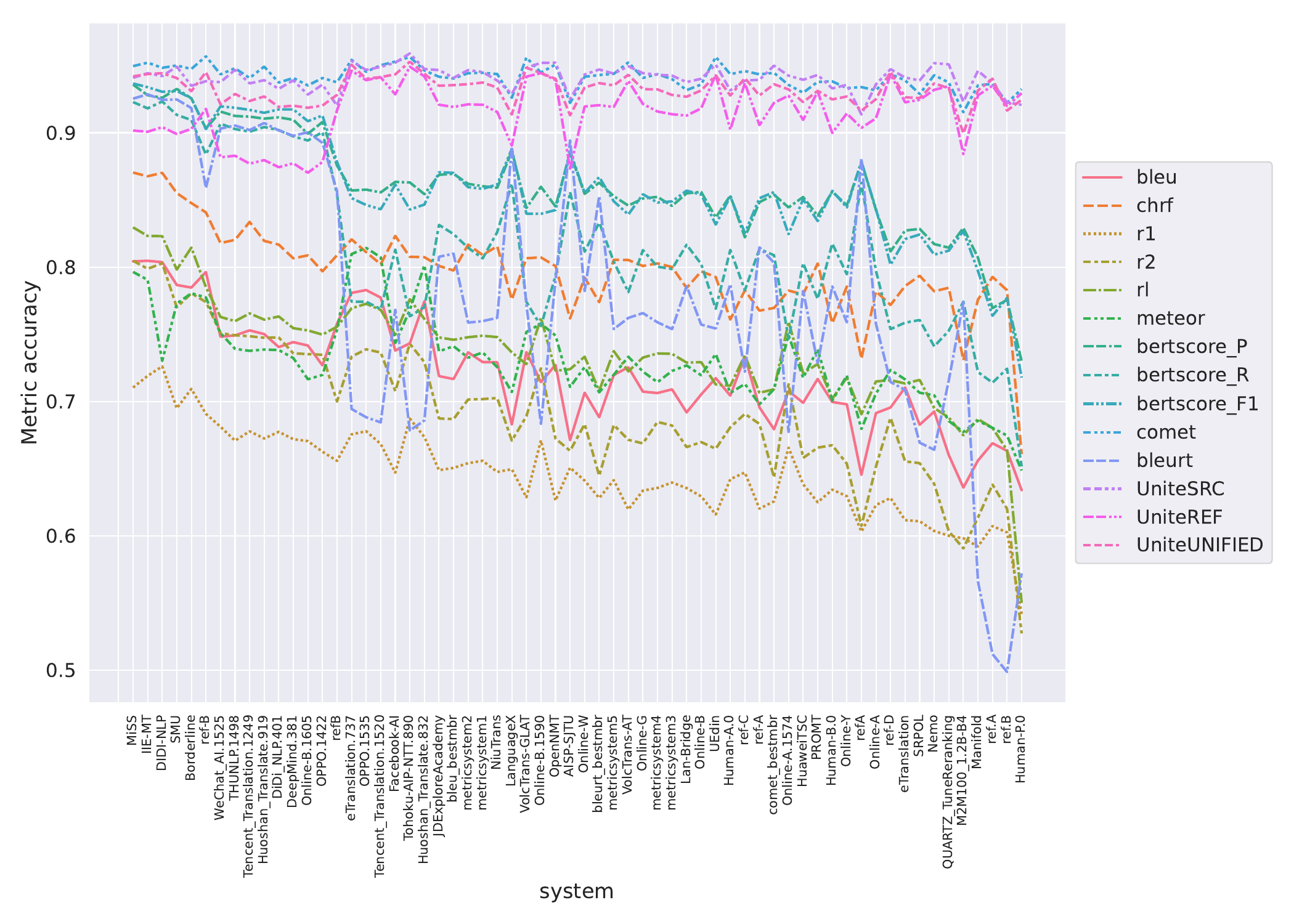}
    \caption{Local metric accuracy across the different contexts}
    \label{fig:mtsystems}
    \end{subfigure}\hfill\begin{subfigure}[t]{0.49\textwidth}
    \centering    
    \includegraphics[width=\textwidth]{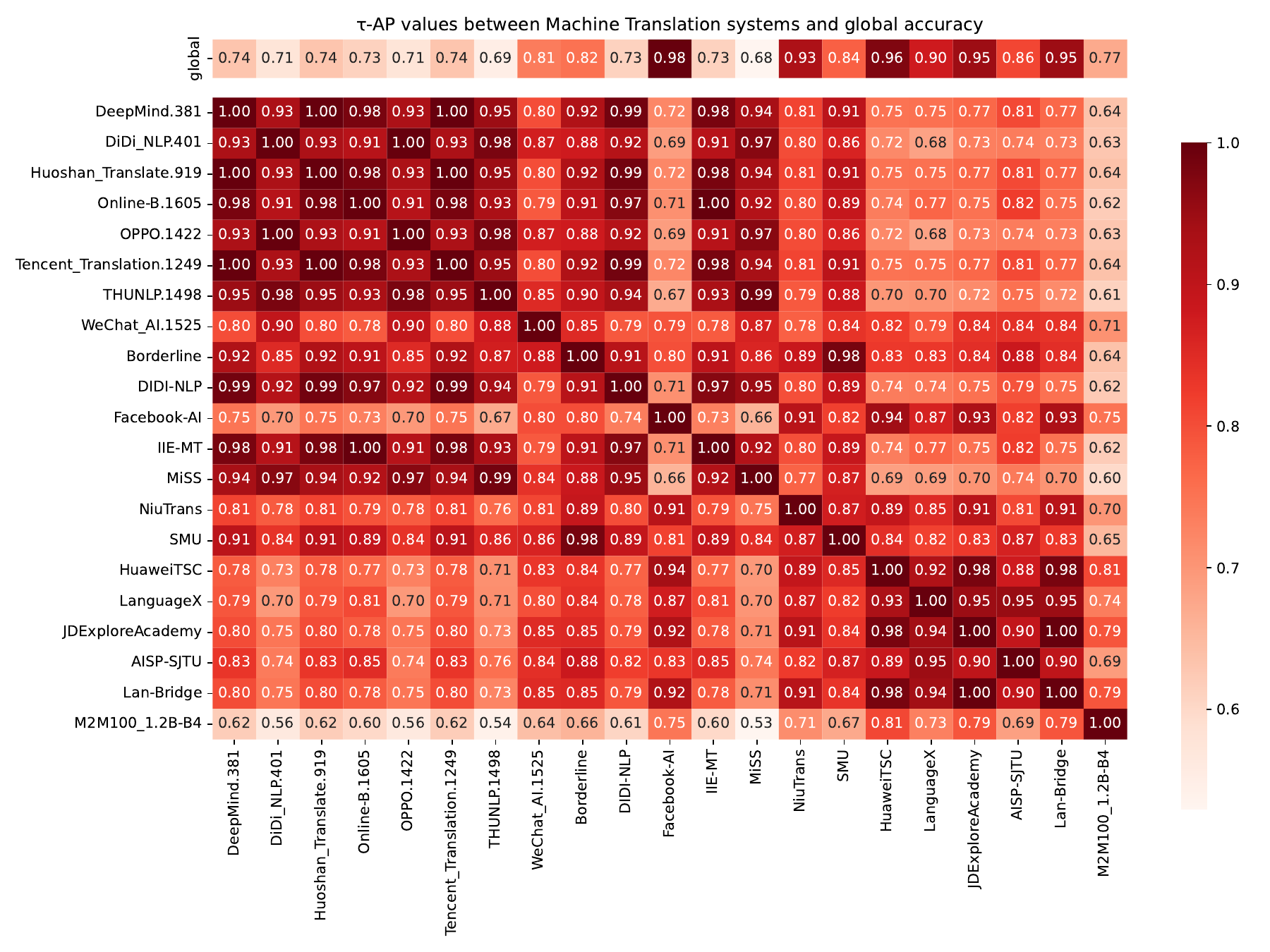}
    \caption{$\tau_{AP}$ values of metrics ordered by local accuracy between different contexts and comparison with global ranking}
    \label{fig:mttau}
    \end{subfigure}
    \caption{Machine Translation.  Metric accuracy for machine translation metrics across the different systems.}
\end{figure*}
\begin{figure*}
    \centering
    \begin{subfigure}[t]{0.49\textwidth}
    \centering    
    \includegraphics[width=\textwidth]{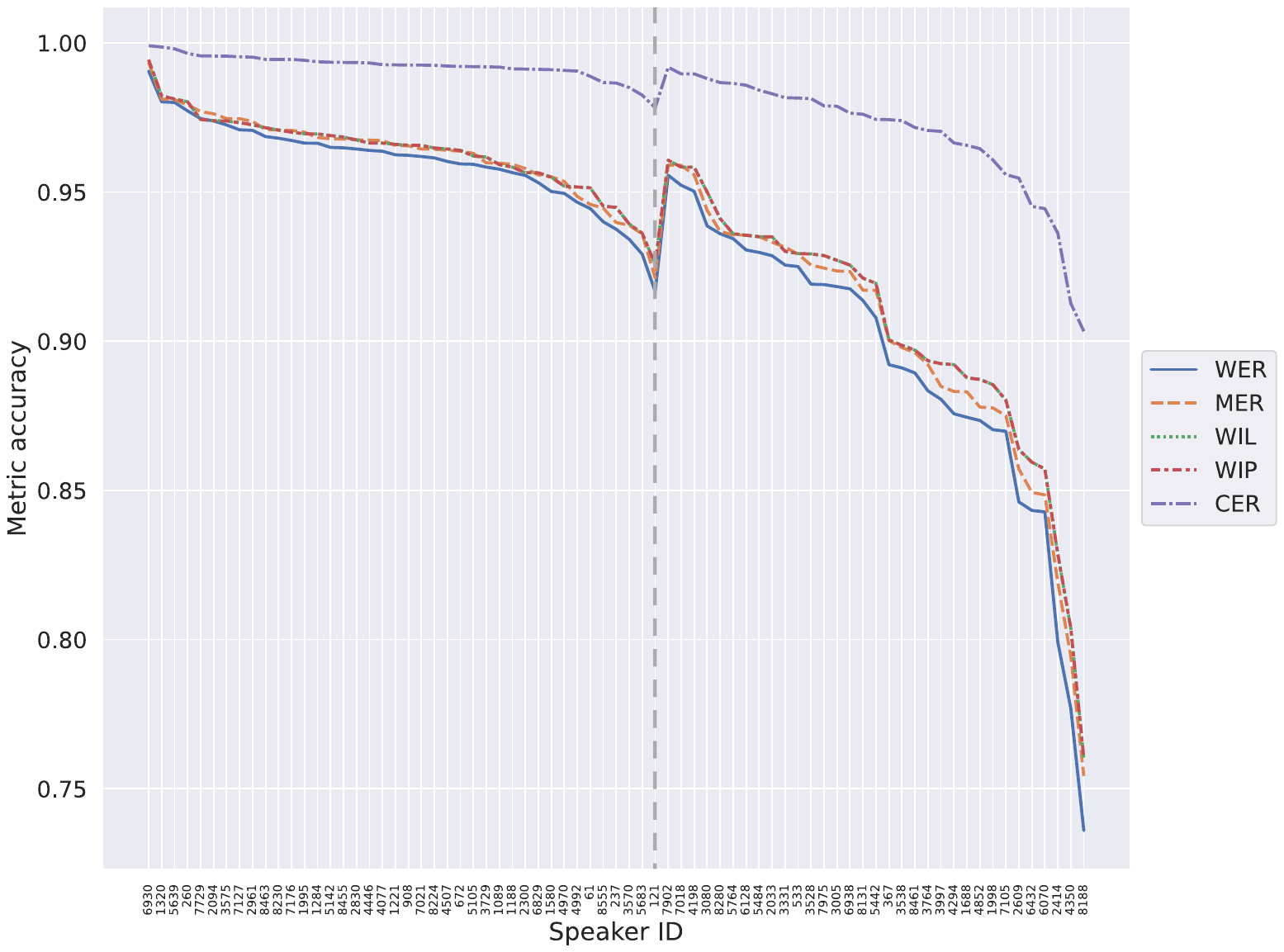}
    \caption{Local metric accuracy across the different contexts}
    \label{fig:asr_speakers}
    \end{subfigure}\hfill\begin{subfigure}[t]{0.49\textwidth}
    \centering    
    \includegraphics[width=\textwidth]{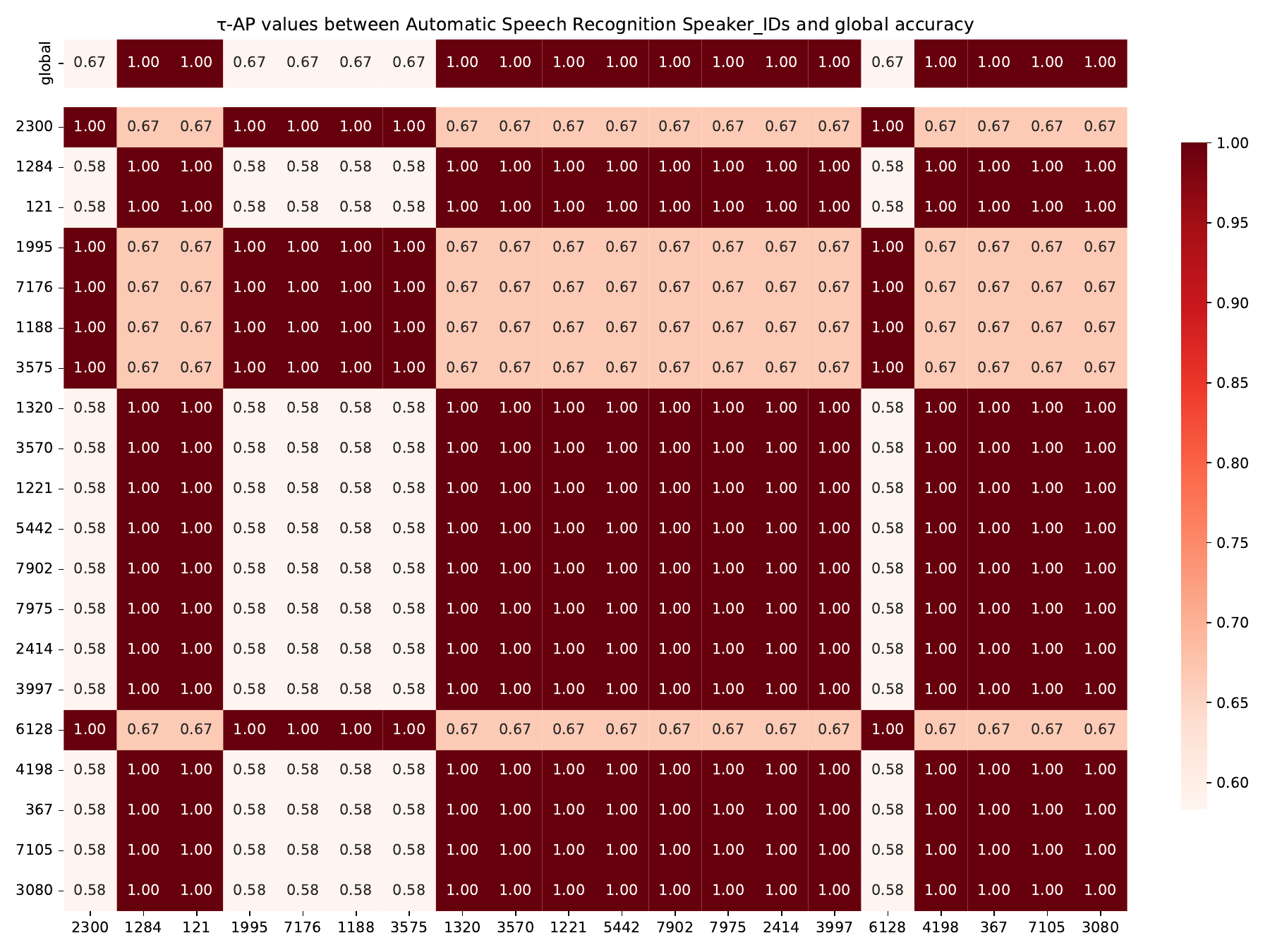}
    \caption{$\tau_{AP}$ values of metrics ordered by local accuracy between different contexts and comparison with global ranking}
    \label{fig:asrtau}
    \end{subfigure}
    \caption{Automatic Speech Recognition. Metric accuracy for automatic speech recognition metrics across the different Speaker IDs. (a) Speaker IDs to the left of the gray line come from the \textsc{Quality=Clean} LibriSpeech-100 dataset, while the Speaker IDs to the right come from the \textsc{Quality=Other} LibriSpeech-100 dataset. }
\end{figure*}

\begin{figure*}
    \centering
    \begin{subfigure}[t]{0.49\textwidth}
    \centering    
    \includegraphics[width=\textwidth]{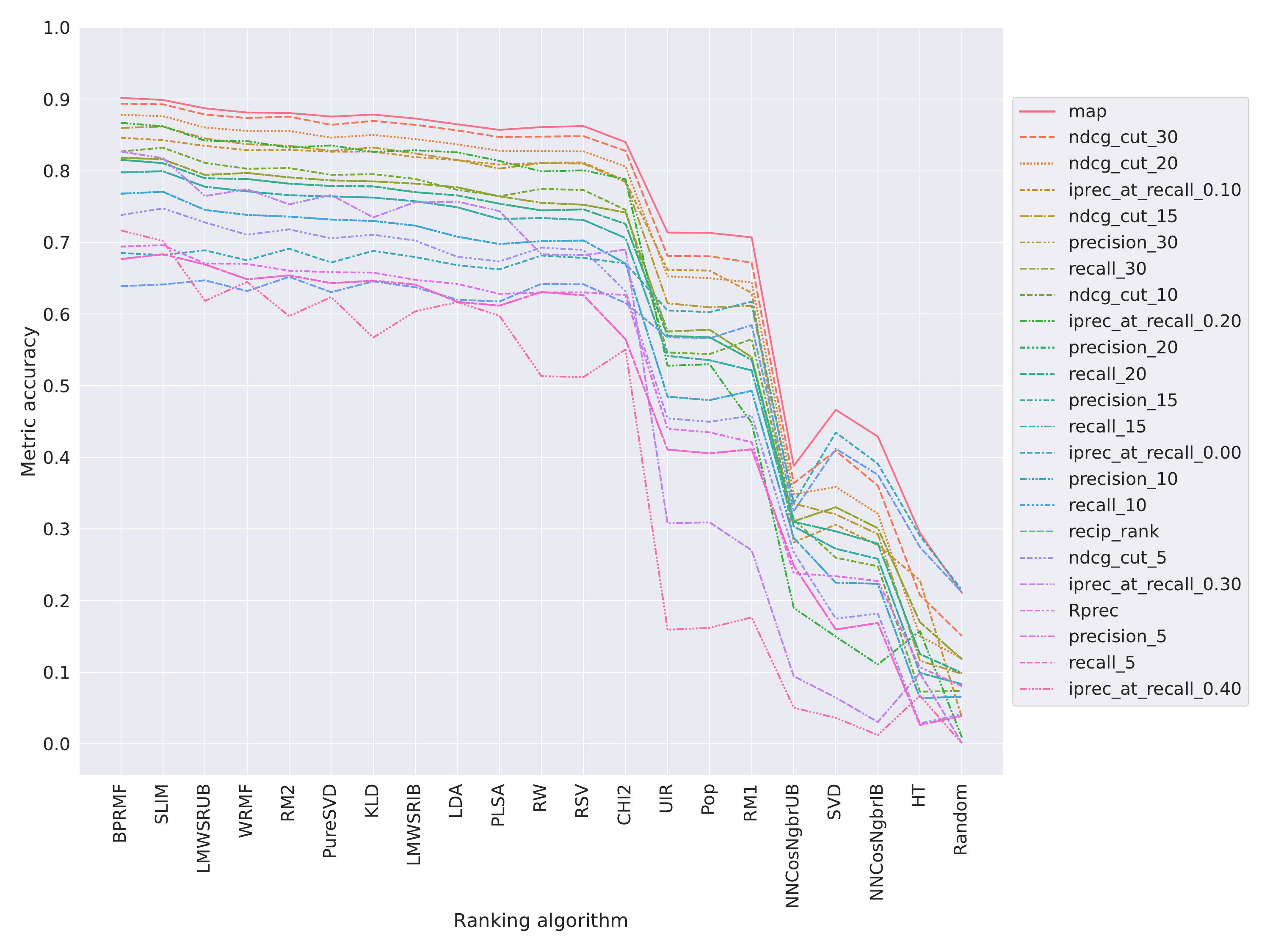}
    \caption{Local metric accuracy across the different contexts}
    \label{fig:ranking_models}
    \end{subfigure}\hfill\begin{subfigure}[t]{0.49\textwidth}
    \centering    
    \includegraphics[width=\textwidth]{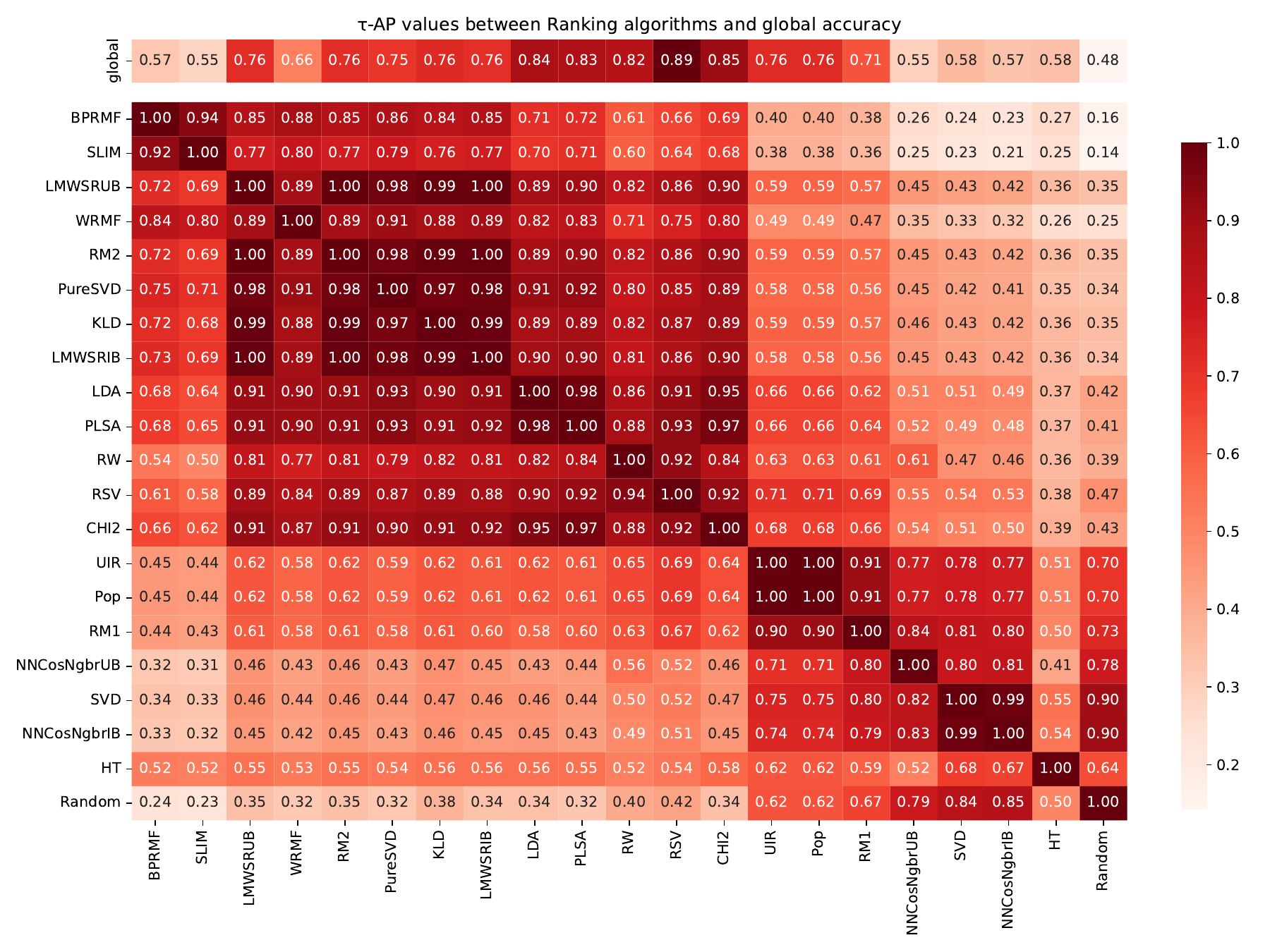}
    \caption{$\tau_{AP}$ values of metrics ordered by local accuracy between different contexts and comparison with global ranking} 
    \label{fig:ranking_kendalltau}
    \end{subfigure}
    \caption{Ranking.  Metric accuracy for ranking metrics across the different systems. }
\end{figure*}

\subsection{Machine Translation}
\label{sec:mt_results}

We visualize the metric accuracies for the machine translation metrics under the different \textsc{System} contexts, as shown in Figure \ref{fig:mtsystems}. We observe that the line for each metric changes as we change the context, as indicated by the varying slopes of the lines. The results of the $\chi^2$ test in Appendix \ref{sec:statsig_results} show that each metric has a $p$-value less than $0.05$, indicating that the difference in the metric accuracies across the different contexts is statistically significant, supporting \textsc{H1} for MT.

Figure \ref{fig:mtsystems} also contains intersections between lines corresponding to different metrics, indicating that there is a change in the relative position of each metric in the different contexts, which also shows a change in the total ordering of metrics by local accuracy across contexts, supporting \textsc{H2}. This is further reinforced in Figure \ref{fig:mttau}, where the $\tau_{AP}$ values show that the correspondence between the pairs of metric accuracy rankings varies considerably for each pair of \textsc{System}s. 

We also visualize the metric accuracies for English to German (\textsc{En-De}) translation pairs under the different \textsc{MQM Score} contexts based on their translation quality \cite{freitag:mqm:tacl} in Figure \ref{fig:mqm_ende}. We provide further details on the scores in Appendix \ref{sec:appendix_mqmfigs}. We observe that the metric accuracies exhibit an almost linear trend across different translation quality contexts: they are low for poor-quality translations and increase as translation quality improves. This pattern suggests that current machine translation metrics struggle to effectively distinguish between very low translation qualities. This aligns with the findings of \citet{fomicheva:mteval:cl}. While metrics are often used to compare state-of-the-art systems with high-quality translations, their ability to differentiate between low-quality outputs is still important. For example, when working with smaller or older models, outputs may be generally of poor quality, yet reliable metrics are still needed to identify promising candidates. We observe similar trends in the graphs for English to Russian (\textsc{En-Ru}) and Chinese to English (\textsc{Zh-En}) translation pairs, which we show in Figures \ref{fig:mqm_enru} and \ref{fig:mqm_zhen} in Appendix \ref{sec:appendix_mqmfigs}, respectively.

\begin{figure}[h]
 \includegraphics[width=0.48\textwidth]{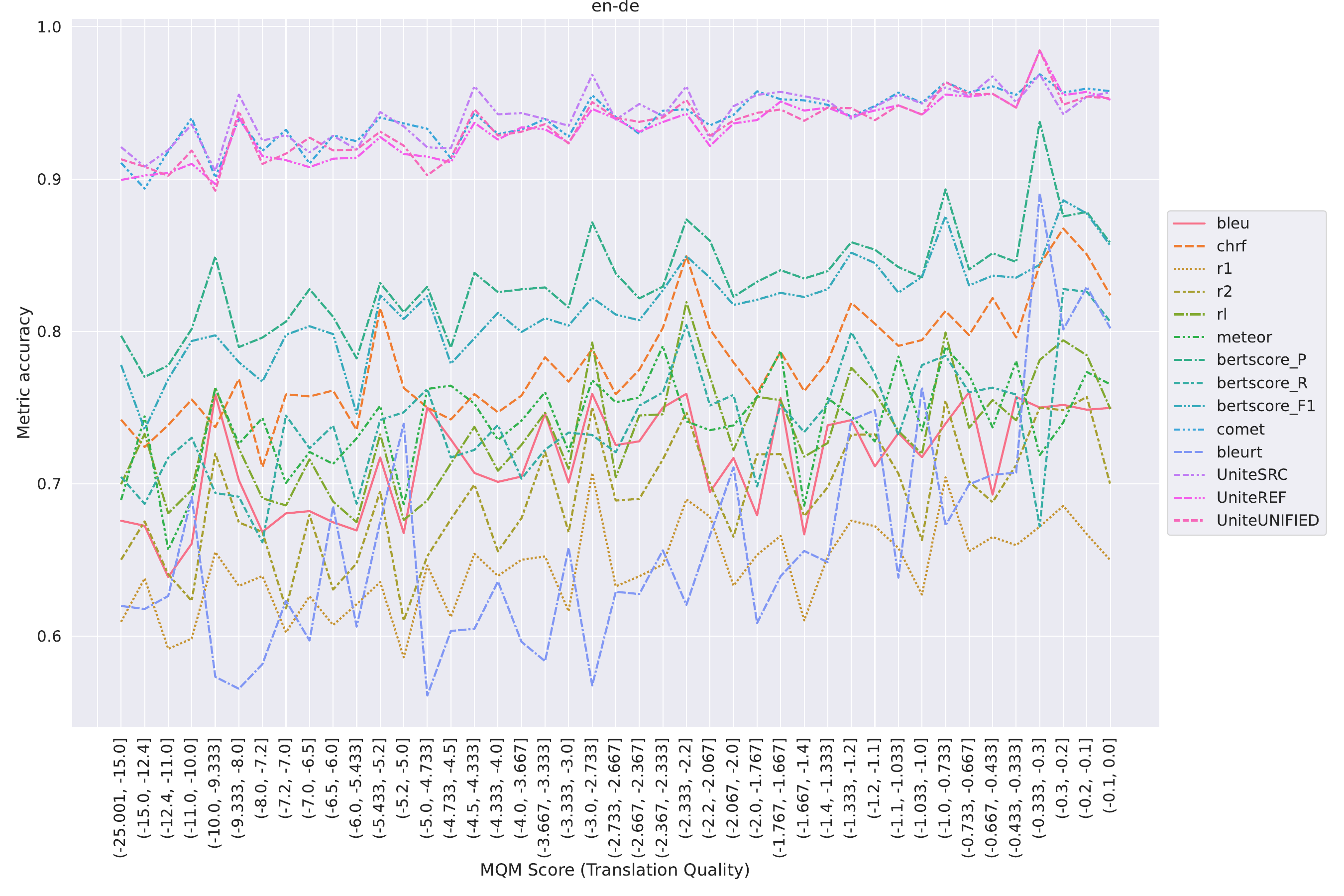}
 \caption{Local metric accuracy across the different MQM scores for English to German (\textsc{En-De}) translation pairs}
 \label{fig:mqm_ende}
\end{figure}

\subsection{Automatic Speech Recognition}
\label{sec:asr_results}

For ASR, we report the local metric accuracy under different \textsc{Speaker ID}s which come from different dataset \textsc{Quality} contexts (\textsc{Clean}/\textsc{Raw}). We plot the local metric accuracy for contexts associated with different contexts shown in Figure \ref{fig:asr_speakers}. We observe that the lines corresponding to each metric are not straight, which indicates that the absolute local accuracy for each metric changes with context, supporting \textsc{H1}. Our $\chi^2$ test results in Appendix \ref{sec:statsig_results} show that each metric has consistent $p$-values less than $0.05$, confirming that the difference in local metric accuracies in the different contexts is statistically significant, providing evidence supporting \textsc{H1} for ASR.

Interestingly, we do not observe the same level of overlaps between the lines corresponding to the different metrics as we did for MT. This, along with the consistently high values of $\tau_{AP}$ for the majority of \textsc{Speaker}s in Figure \ref{fig:asrtau}, indicates that there is no evidence to support \textsc{H2} for ASR. We will discuss this observation in more detail in \ref{sec:reladisc}.

\subsection{Ranking}
\label{sec:ranking_results}

Putting the metric accuracies for the ranking metrics for the different \textsc{Algorithm}s in Figure \ref{fig:ranking_models}, we can first observe that none of the lines corresponding to the different metrics are straight lines, which supports \textsc{H1}. The large fluctuations within each line suggest that the changes in the absolute local accuracies for each metric are rather significant. The $\chi^2$ test in Appendix \ref{sec:statsig_results} shows that each metric has a consistent $p$-value less than $0.05$, which means a statistically significant change, providing evidence to support \textsc{H1}.

Furthermore, we can see that overlaps exist between the different lines corresponding to the different metrics, similar to the observation we made in the MT case. This indicates that the total ordering of metric accuracies changes as the context changes, supporting \textsc{H2}. The $\tau_{AP}$ results (Figure \ref{fig:ranking_kendalltau}) show clustering by algorithm, as with MT.

\section{Discussion}
\label{sec:discussion}

\subsection{\textsc{H1}: Absolute Local Accuracies}
\label{sec:aladisc}

The results in Section \ref{sec:results} generally provide evidence supporting \textsc{H1}, as our experiments consistently show that the local metric accuracy changes as the context changes. 

Our perturbation method catastrophically degrades good outputs, and bad outputs are already poor and difficult to make demonstrably worse. In Figure \ref{fig:ranking_models}, we can observe that the local metric accuracy for a context is related to the average quality of outputs in that context. For example, in the ranking setting, the most effective system according to MAP is BPRMF, which is also the context whose perturbed outputs are easiest to distinguish. In contrast, perturbed outputs in the random ranker are more difficult for all metrics to distinguish. We will return to this in Section \ref{sec:methoddisc}. 

The results corresponding to how the local metric accuracy changes as the context changes suggest that evaluators may be interested in the stability of local metric accuracy when selecting a metric. A more stable metric is more predictable when deployed in a new context, and the probability of selecting the wrong system remains consistent. This is especially important if we consider a new evaluation context where poor local metric accuracy puts users---or a vulnerable subgroup of users---at risk. Work in robust machine learning provides existing methods for designing metrics stable across context changes \cite{yuan:robustness:neurips}.

In addition to stability, we can organize metrics according to systematic behavior in the local metric accuracy. For example, in Figure \ref{fig:mtsystems}, more complex embedding-based and model-based evaluation metrics generally perform better than the simpler lexical-based metrics \cite{zhang:bertscore:iclr, freitag:wmt2022:wmt}. More complex metrics cluster at the top of the figure, while simpler metrics occupy the bottom regions; any overlap occurs mostly within a specific subset. Such an analysis allows evaluators to understand the empirical relationships between metric ensembles. Although picking the best metric might involve selecting a metric that occupies the top of the figure, there may be contexts in which local metric accuracies are close enough to allow flexibility in selecting metrics with lower local metric accuracy. 

More generally, we can consider multi-objective metric development. For example, since embedding- and model-based methods are more time-intensive and computationally costly compared to the lexical-based methods, adopting simpler and cheaper metrics when local metric accuracies are comparable (e.g., early in model development) would result in cost savings and faster iteration. Beyond cost and local metric accuracy, one can imagine local versions of metric interpretability, metric engineering overhead, metric optimizability, and other criteria when conducting contextual meta-evaluation.

\subsection{\textsc{H2}: Relative Local Accuracies}
\label{sec:reladisc}

Although the observations in Section \ref{sec:results} err toward accepting \textsc{H2}, the evidence from our experiments on the ASR task in Section \ref{sec:asr_results} suggests that it depends on the task and their corresponding metrics. In ASR, the ambiguity of the correct answers is low, unlike in MT or ranking, where two outputs (e.g., translations or permutations, respectively) can be equally valid \cite{wieting:beyondbleu:acl}. ASR outputs are typically direct transcriptions of spoken language into text, and there is often a single `correct' or `expected' output for a given input. The main evaluation objective is to check whether the transcription accurately matches the spoken words without considering other criteria such as grammatical structure or fluency. Furthermore, ASR tasks rely on phonetic accuracy, and the primary goal is to replicate spoken words as text as accurately as possible. Although there may be challenges that make the task difficult--such as the presence of accents, homophones, and background noise--evaluating the task itself does not require deep semantic understanding beyond recognizing the correct phonetic and lexical forms, unlike machine translation. Hence, the metrics commonly used to benchmark ASR systems only slightly vary in the characteristics they are trying to measure, and they are all operationalized following similar statistical methods.

Figures \ref{fig:mttau} and \ref{fig:ranking_kendalltau} indicate that there are groups of contexts where the relative reliability of the metrics is similar (i.e., the relative ranking of metrics in these groups of contexts does not change significantly). When contexts can be structured according to metric accuracy, one can adopt a fixed evaluation metric. This has practical implications in terms of engineering and development overhead or, in the case of model-based metrics, model development cost. \textit{Predicting} the similarity in local metric accuracy ordering (i.e., cells in Figures \ref{fig:mttau} and \ref{fig:ranking_kendalltau}) is an important task because it allows evaluators to confidently adopt an evaluation metric without conducting contextual meta-evaluation. Predictive features include any metadata we have about the contexts. For example, in ranking, \citet{valcarce:topn:acmrecsys} categorize \textsc{Algorithm}s into different families of techniques: matrix factorization (\textsc{SVD}, \textsc{PureSVD}, \textsc{BPRMF}, \textsc{WRMF}), neighborhood-based (\textsc{CHI2}, \textsc{KLD}, \textsc{RSV}, \textsc{Rocchio's Weights}).

Additionally, the top rows on Figures \ref{fig:mttau} and \ref{fig:ranking_kendalltau} show that the ranking of the metrics in different contexts varies considerably from the ranking of the metrics based on the global accuracy. This observation aligns with the scenario illustrated by the toy example in Table \ref{tab:toy_example}, where we show that selecting the most appropriate metric may not be as straightforward as selecting the metric with the highest global precision and further motivates the need to shift towards context-specific metric evaluations.

\subsection{Methodology}
\label{sec:methoddisc}

Although our results demonstrate that local metric accuracy analysis can provide insight into metric behavior, there are several opportunities to improve the methodology. First, our perturbations, while reliable in generating output degradations, may result in outputs that are easily detected by metrics, especially for highly effective systems. Moreover, perturbed outputs may be sufficiently different as to be unlikely to occur in a specific context. For example, if we are evaluating in the context of highly effective MT systems, a translation with a missing word is very unlikely by any highly effective MT system, even though we know that it is of lower quality. In order to address this, developing perturbation methods that reliably degrade performance \textit{and} are likely to occur within a context will be important for future local metric accuracy development. This is related to synthesizing hard negative examples in the contrastive learning literature \cite{kalantidis:hard-negative-mixing}. Alternatively, we can consider non-perturbation data, perhaps from human annotators, although this compromises the cost-effectiveness of output perturbation.

Finally, in order to help with clarity, we focused on contexts that were interpretable and which contexts were relevant depending on the broader model evaluation environment. Focusing on models, as we did for MT and ranking, emphasizes contexts that reflect iterative model development and refinement within a narrow set of constraints (i.e., the particular model being evaluated). If we benchmark a diverse set of systems, we are interested in comparing a broader set of possible outputs than those from a single system. In cases where we are designing a metric agnostic to a particular context, we may be interested in robust performance across arbitrary contexts. Although this is similar to global analysis, a more rigorous and formal approach to context selection, such as found in the distributionally robust machine learning literature \cite{duchi:robust-optimization}, may be more appropriate.

\subsection{Practical guidelines}
\label{sec:practicalguidelines}
Our method serves as a diagnostic tool for metric meta-evaluation, similar to how practitioners use ablation studies for error analysis. Our findings from \textsc{H1} on varying absolute accuracy suggest that evaluators should examine how consistently metrics perform within their context rather than relying on global measures. For example, when developing early-stage models, focus on metrics that show stable performance with low-quality outputs. Our findings from \textsc{H2} on varying relative ordering indicate that metric selection should be context-dependent, which means that a metric that performs well for comparing high-quality models might not be optimal for comparing early-stage models.

Consider a scenario where a team is developing a machine translation model for a low-resource language pair. Early in the development cycle, the model produces outputs that are far from the target distribution, often containing grammatical errors and incorrect translations. The team applies our methodology to evaluate metrics in the context of outputs with varying quality. They find that simpler lexical-based metrics, such as BLEU, show more stable performance in distinguishing between poor translations, while more complex metrics like COMET struggle to differentiate between outputs of similarly low quality but can differentiate between outputs of similarly high quality. Based on this insight, the team selects BLEU for early-stage evaluation, ensuring they can reliably track incremental improvements. As the model develops and outputs improve, the team would adapt and switch to more sophisticated metrics like COMET to evaluate high-quality translations. This example demonstrates how our method can guide metric selection at different stages of model development, ensuring that the chosen metrics align with the specific evaluation context.

To summarize, when using our approach, evaluators can follow these steps:

\begin{enumerate}
    \item Identify the specific context: define the evaluation context, such as the type of system (e.g., early-stage vs. late-stage models), data domain (e.g., medical vs. legal text), etc.
    \item Measure local metric accuracies: use our existing results or apply our method to measure how well different metrics distinguish quality differences within that context.
    \item Select metrics based on stability and context: choose metrics that demonstrate stable accuracies for the specific use case. For example, if the context involves low-quality outputs, prioritize metrics that reliably differentiate between poor results. If the context involves high-quality outputs, select metrics that are sensitive to more subtle or fine-grained differences.
    \item Monitor and adapt metrics: as the evaluation setting changes (e.g., the model is iteratively trained), regularly reassess the performance of the selected metrics to ensure they are still appropriate. For example, transition from simpler lexical-based metrics for early-stage models to more sophisticated embedding-based metrics for late-stage models.
\end{enumerate}

By following these steps, practitioners can make informed decisions about metric selection, ensuring that their evaluations are contextually appropriate. This approach not only improves the reliability of model benchmarking but also reduces the risk of selecting suboptimal metrics that may lead to incorrect conclusions about system performance.

\section{Conclusion}

We introduce the notion of local metric accuracy and demonstrate how to use it to conduct contextual metric meta-evaluation. Our results show that both the absolute and relative local accuracy of a metric varies across evaluation contexts, although the extent of variation depends on the task. Our findings highlight the importance of moving beyond global metric meta-evaluation to better understand metric performance. This will allow practitioners to make more informed decisions, leading to more reliable evaluations and reducing the risk of selecting suboptimal metrics.


\section{Limitations}

As mentioned in Section \ref{sec:perturbation_techniques} and Appendix \ref{sec:perturbation_implementation}, our experiments adopt relatively simple rule-based and LLM-based perturbation methods to cover several tasks. However, given the wide range of system output qualities in our dataset, it is rather difficult to guarantee that applying a perturbation function to $y$ will always produce a $y'$ that is worse in utility.

We also compute local accuracy by uniformly weighting all output-perturbation pairs, which may not reflect the true distribution of outputs in a specific context. In reality, different outputs have different probabilities of occurring in a specific context. These probabilities should be incorporated into the accuracy calculation to provide a more reliable estimate of local metric accuracy. Estimating the distribution over outputs for a specific context itself is a difficult research question that we plan to address in future work.

\section{Acknowledgements}

We would like to thank Ben Carterette, Alon Lavie, Graham Neubig, Emine Yilmaz, Alfredo Gomez, Shaily Bhatt, To Eun Kim, Jessica Huynh, Dennis Frauen, and Lars van der Laan for their valuable feedback throughout the progress of this work. Special thanks to Shinji Watanabe and Yifan Peng for the ASR outputs used in our experiments.

\bibliography{custom}

\appendix

\section{Appendix}
\label{sec:appendix}
\subsection{Details for Tasks, Datasets, and Metrics}
\label{sec:materials_details}

\begin{table*}[t]
\centering
{\small
\begin{tabular}{lp{2.7in}p{2.7in}}

\textbf{Task} & \textbf{Dataset} & \textbf{Metrics} \\
\hline
\\
MT & Over 150,000 system outputs and reference translation from 62 different MT systems 
 submitted to the WMT metrics task from the year 2022 and prior \cite{freitag:wmt2022:wmt} with MQM annotations \cite{freitag:mqm:tacl} for the source-target language pairs English-Russian (\textsc{En-Ru}), English-German (\textsc{En-De}), Chinese-English (\textsc{Zh-En}). The subsets that are available are \textsc{Year}, \textsc{Domain}, \textsc{System}, and \textsc{MQM Score}. The dataset is available on HuggingFace\footnote{\url{https://huggingface.co/datasets/RicardoRei/wmt-mqm-human-evaluation}}. & \textsc{Bleu}, \textsc{Rouge-1}, \textsc{Rouge-2}, \textsc{Rouge-L}, \textsc{Meteor}, \textsc{BertScoreP}, \textsc{BertScoreR}, \textsc{BertScoreF1}, \textsc{Comet}, \textsc{BleuRT}, \textsc{ChrF}, \textsc{UniteSRC}, \textsc{UniteREF}, \textsc{UniteUNIFIED} \\\\
 ASR & Over 33,000 system outputs from six different ASR models on ESPnet \cite{watanable:espnet:interspeech} on the LibriSpeech 100 dataset \cite{panayotov:librispeech:icassp}. The subsets that are available are \textsc{System}, \textsc{Speaker ID}, and \textsc{Quality}. & Word Error Rate (\textsc{WER}), Match Error Rate (\textsc{MER}), Word Information Lost (\textsc{WIL}), Word Information Preserved (\textsc{WIP}), Character Error Rate (\textsc{CER}) \\\\
Ranking & Ranked list of top-100 items retrieved by 21 recommender algorithms provided by \citet{valcarce:topn:acmrecsys} on the MovieLens1M dataset \cite{maxwell:movielens:acm} submitted to TREC \cite{buckley:trec:acmsigir}. We were able to segment the outputs by \textsc{Algorithm}.
& Mean Average Precision (\textsc{MAP}), Precision@$R$, where $R$ is the number of relevant documents (\textsc{Rprec}), Reciprocal Rank (\textsc{Recip\_Rank}), Interpolated Precision at Recall Level $X$ (for $X = \{0.0, 0.1, 0.2, 0.3, 0.4\}$) (\textsc{iprec\_at\_recall\_X}), Precision@$K$ (\textsc{P\_K}), Recall@$K$ (\textsc{Recall\_K}), nDCG@$K$ (\textsc{nDCG\_cut\_K}) (where $K={5, 10, 15, 20, 30}$)
\end{tabular}
}
\caption{Datasets and metrics used for different tasks}
\label{tab:datasets_metrics}
\end{table*}

Table \ref{tab:datasets_metrics} shows the dataset and metrics that we used in our experiments. For each task, we used readily available system outputs to improve reproducibility. For each metric, we employ its respective official implementations or, when unavailable, the most widely used implementation with default parameters. For any neural metric computation, we used an NVIDIA RTX A6000 GPU. For \textsc{BLEU}, we used \texttt{sentencebleu} implementation from \texttt{nltk}\footnote{\url{https://www.nltk.org/index.html}}. We also used \texttt{nltk}'s implementation for \textsc{Meteor}. For \textsc{Rouge}\footnote{\url{https://github.com/google-research/google-research/tree/master/rouge}} and \textsc{BertScore}\footnote{\url{https://github.com/Tiiiger/bert_score}}, we have used the implementation released by their respective authors. For \textsc{BleuRT}\footnote{\url{https://github.com/google-research/bleurt\#readme}}, \textsc{Comet}\footnote{\url{https://unbabel.github.io/COMET/html/models.html}}, \textsc{ChrF}\footnote{\url{https://github.com/mjpost/sacreBLEU\#chrf--chrf}} and \textsc{Unite}\footnote{\url{https://huggingface.co/Unbabel/unite-mup}}, we have used their official implementations via the \texttt{evaluate} library on HuggingFace. We have used the default\footnote{\url{https://huggingface.co/Unbabel/wmt22-comet-da}} model for \textsc{Comet}.
We used the \texttt{jiwer}\footnote{\url{https://github.com/jitsi/jiwer}} Python package to compute the ASR metrics. We used the \texttt{trec\_eval}\footnote{\url{https://github.com/usnistgov/trec_eval}} to calculate the ranking metrics.

\subsection{Perturbations}

\subsubsection{List of perturbations}
\label{sec:perturbation_appendix}
The complete list of perturbations used in our experiments and their descriptions are in Table \ref{table:perturbations}.

\subsubsection{Implementation of perturbations}
\label{sec:perturbation_implementation}
For the system output corresponding to machine translation and automatic speech recognition tasks, we first preprocess the target output by performing tokenization, part-of-speech tagging, named entity recognition, negation detection, etc. Additionally, the WMT dataset \cite{freitag:wmt2022:wmt} contains target outputs in three different languages--English, Russian, and German--we were careful to use libraries that are tailored to each language. We then use simple string manipulation functions to perform rule-based insertion, deletion, substation, and more. To introduce an additional layer of complexity, we also prompted an instruction-tuned LLM to perturb the system and generate a degraded output. 

We employ simple rule-based perturbation techniques with the addition of an LLM-based perturbation. We noted that not all perturbation techniques are applicable to each task, and they are also not applicable to each system output in each task. The applicability here is defined by whether or not applying the perturbation function $f(\cdot)$ to $y$ will produce a $y'$ that is worse with high probability.

\textbf{Swapping. }We apply this perturbation method to the outputs of all three tasks. We tokenize lowercased sentences for machine translation and automatic speech recognition using \texttt{nltk}'s \texttt{word\_tokenzize} function. We apply this preprocessing method to all the rule-based perturbation functions. Then, we randomly pick a pair of words and swap their positions. Similarly, for ranking, we pick a pair of items within the ranked list and swap them. To offset the fact that we have a limited number of perturbations for ranking, we swap several items within the list, while in machine translation and automatic speech recognition, we only switch the position of a pair of words.

\textbf{Removal. }We apply this perturbation method to outputs from machine translation and automatic speech recognition. After preprocessing the output, we randomly select a word from the sentence and remove it.

\textbf{Insertion. }We apply this perturbation method to outputs from machine translation and automatic speech recognition. After preprocessing the output, we insert a randomly selected word from a list of words corresponding to the language of the system output. To reduce computation time, we store a corpus of English, Russian, and German words from the dataset and randomly select one from the list.

\textbf{Synonym substitution. }We apply this perturbation method to the output of machine translation. We have to take a language-specific approach to obtain synonyms: for English, we used \texttt{nltk wordnet}'s \texttt{synset}s; for Russian, we used \texttt{synset}s from \texttt{RuWordNet}\footnote{\url{https://www.ruwordnet.ru/ru}}; for German, we performed an API call using the \texttt{PyMultiDictionary}\footnote{\url{https://github.com/ppizarror/PyMultiDictionary}} library. Not every word has a synonym, so we iterate through the sentence and substitute the first word with a synonym.

\textbf{Pronoun substitution. }We apply this perturbation method to the output of machine translation. We adopted a strictly rule-based approach, defining a set of pronouns for each language (English, Russian, and German). We loop through the words in the sentence, look at all available pronouns, and replace one with a random pronoun from the predefined list.

\textbf{Named entity substitution. }We apply this perturbation method to the output of machine translation. Similar to pronoun substitution, we adopted a strictly rule-based approach, where for each language, we collect a list of named entities by performing named entity recognition on all the sentences using \texttt{spacy}\footnote{\url{https://spacy.io/}}. Then, for each sentence from the system output, we look for named entities using the same approach and replace them with a randomly named entity from our cached list of named entities corresponding to the language.

\textbf{Negation. }We apply this perturbation method to the output of machine translation. For English outputs, we perform POS tagging using \texttt{nltk}. Whenever we see a verb, adjective, or auxiliary verb, we insert a `not' in the appropriate position (before the word in the first two cases and after the word in the latter). For German, instead of performing POS tagging, we defined a list of common auxiliary verbs, adjectives, and verbs and then added a `nicht' in the appropriate position. We do the same thing for the Russian outputs.

\textbf{Negation removal. }We apply this perturbation method to the output of machine translation. Semantically speaking, negating a sentence is the same as removing a negation (for example, removing the word `not' from a sentence results from adding a double negation to a sentence). However, we operationalize this differently in the implementation. We adopt a strictly rule-based approach where, for each language, we look for instances of negation like ``not'', ``didn't'', ``never'', and replace them with their negated counterpart (or for the case of words like ``never'', just removed them entirely).

\textbf{Phonetic substitution. }We apply this perturbation method to outputs from automatic speech recognition. Like several methods described above, we adopted a rule-based method to replace consonants and vowels (or consonant and vowel groups) with something phonetically similar. For example, we can perform a \texttt{t -> d} substitution to \texttt{water -> wader}. We do this using \texttt{regex}.

\textbf{LLM-based perturbation. }We apply this perturbation method to outputs from machine translation and automatic speech recognition. We employ a \texttt{Llama-3-8B-instruct} model \cite{llama3modelcard} to perturb a system output. We use the following hyperparameters: \texttt{temperature=0.3}, \texttt{max\_response\_tokens=200}, and \texttt{top\_p=1.0}. We run inference on a single NVIDIA A6000 GPU. Below is an example prompt that we use to perturb a machine translation system output:

\begin{quote}
``I have the following source text (SOURCE\_ENGLISH), reference translation (REFERENCE\_GERMAN), and machine translation system output (MT\_OUTPUT\_GERMAN). Your task is to make MT\_OUTPUT\_GERMAN slightly worse by introducing minor machine translation errors while keeping the output generally understandable. 

SOURCE\_ENGLISH: ``Fiddes also addressed allegations of child sexual abuse against Jackson by Wade Robson and James Safechuck, which were aired in the controversial, Emmy Award-winning documentary Leaving Neverland.''

REFERENCE\_GERMAN: ``Fiddes hat auch die Anschuldigungen des sexuellen Kindermissbrauchs von Wade Robson und James Safechuck durch Jackson angesprochen, die in der kontroversen Dokumentation „Leaving Neverland'' die auch einen Emmy gewonnen hat, dargelegt wurden.''

MT\_OUTPUT\_GERMAN: ``Fiddes befasste sich auch mit Vorwürfen des sexuellen Missbrauchs von Kindern gegen Jackson durch Wade Robson und James Safechuck, die in dem umstrittenen, Emmy-prämierten Dokumentarfilm Leaving Neverland ausgestrahlt wurden.''

Please modify MT\_OUTPUT\_GERMAN to introduce minor typical machine translation errors, making the output only slightly worse. Do not include any explanations or additional text. Only return the perturbed output.''	
\end{quote}

The prompt above yields the following output:

\begin{quote}
``Fiddes befasste sich auch mit Vorwürfen des sexuellen Missbrauchs von Kindern gegen Jackson von Wade Robson und James Safechuck, die in dem umstrittenen, Emmy-premierten Dokumentarfilm Leaving Neverland ausgestrahlt wurden.''
\end{quote}

Similarly, we present an example prompt that we use to perturb an automatic speech recognition system output:

\begin{quote}
I have the following reference text (ASR\_REFERENCE\_TEXT) and an ASR system output (ASR\_SYSTEM\_OUTPUT). Your task is to make ASR\_SYSTEM\_OUTPUT slightly worse by introducing minor ASR errors.

ASR\_REFERENCE\_TEXT: ``concord returned to its place amidst the tents''
ASR\_SYSTEM\_OUTPUT: ``concord returned to its place amidst the tents''

Please modify ASR\_SYSTEM\_OUTPUT to introduce minor typical ASR errors, making the output only slightly worse. Do not include any explanations or additional text. Only return the perturbed output.
\end{quote}

The prompt above yields the following output:

\begin{quote}
 ``concord reterned to its pace amids the tents''
\end{quote}

We use the same prompt template for all system outputs used for the meta-evaluation. In both cases, we do not guide the model by using examples such as ``perturb the model by dropping some words'' or other detailed instructions. This will allow the LLM to `explore' different ways to perturb an output beyond our rule-based methods.

\begingroup
\setlength{\tabcolsep}{6pt} 
\renewcommand{\arraystretch}{1.5} 
\begin{table*}[]
{\small
\centering
\begin{tabular}{lp{0.9in}p{4.5in}}
\textbf{Task} & \textbf{Perturbation} & \textbf{Description} \\ \hline
\multirow{9}{*}{MT} 
 & Removal & Drop a random word from the string, e.g. ``John said he did not like the cake baked.'' \\ 
 & Insertion & Add a random word to the string at a random position, e.g. ``John said he did not \colorbox{pink}{really} like the cake she baked.'' \\ 
 & Swapping & Switch the position of a random pair of words from the string, e.g.``John \colorbox{pink}{he} \colorbox{pink}{said} did not like the cake she baked.'' \\
 & Synonym sub. & Substitute a random word with its synonym, e.g. ``John said he did not \colorbox{pink}{enjoy} the cake she baked.'' \\ 
 & Pronoun sub. & Substitute a random pronoun in the string with another random pronoun, e.g. ``John said \colorbox{pink}{she} did not like the cake she baked''\\
 & Named entity sub. & Substitute a random named entity in the string with another named entity, e.g. ``\colorbox{pink}{Mike} said he did not like the cake she baked.'' \\ 
 & Negation & Negate a random word in the string, e.g. ``John said he did not like the cake she \colorbox{pink}{did not} baked.'' \\ 
 & Negation removal & Remove instances of negation from the string, e.g. ``John said he liked the cake she baked.'' \\ 
 & LLM & Prompt an LLM to perturb the string, e.g. ``John \colorbox{pink}{mentioned} he \colorbox{pink}{didn’t} \colorbox{pink}{enjoy} the cake she \colorbox{pink}{made}.''\\ \hline
\multirow{5}{*}{ASR} 
 & Removal & Drop a random word from the string, e.g. ``John said he did not like cake she baked.'' \\ 
 & Insertion & Add a random word to the string at a random position, e.g. ``John said he did not like the cake \colorbox{pink}{that} she baked.'' \\ 
 & Swapping & Switch the position of a random pair of words from the string, e.g. ``John said he did \colorbox{pink}{like} \colorbox{pink}{not} the cake she baked.'' \\ 
 & Phonetic sub. & Replace a random letter (or a letter group) with phonetically similar ones, e.g. ``John \colorbox{pink}{sed} he did not like the cake she \colorbox{pink}{bakked}.'' \\ 
 & LLM & Prompt an LLM to perturb the string, e.g. ``John \colorbox{pink}{sed} he did not like cake she baked.'' \\ \hline
\multirow{1}{*}{Ranking} 
 & Swapping & Switch the position of an item with another item within the ranked list, e.g. Original ranking: [Cake, Pie, Cookie] Perturbed: [Cookie, Pie, Cake]\\ \hline
\end{tabular}}
\caption{Full list of perturbations for MT, ASR, and Ranking tasks. The original unperturbed sentence for the MT and ASR examples is ``John said he did not like the cake she baked.'' We outline how each perturbation method is implemented in Appendix \ref{sec:perturbation_implementation}.
}
\label{table:perturbations}
\end{table*}
\endgroup

\subsubsection{The effect of adding more perturbations}
\label{sec:more_perturbations_good}
In this section, we will discuss how we can reduce the effects of potential confounding between the choice of perturbation and the different metrics being evaluated. We can detect whether a metric confounds with a specific perturbation method by observing its variance across the different perturbation methods, as initially highlighted in Section \ref{sec:perturbation_techniques}.

We can think about how an adversarial metric can take advantage of such confounders and how introducing more perturbations would help. Let us say we only adopt the removal perturbation function in our framework. Let $\mu_{adv}$ be the adversarial metric that assigns a score based on the length of the system output $y$, such that $\mu_{adv}(x,y) = len(y)$. Given that $y'$ will always be shorter than $y$ due to the perturbation function that we chose, it is always true that $\mu_{\text{adv}}(x,y')$ is always lower than $\mu_{\text{adv}}(x,y)$, hence we will have $ACC_\mu(Q_x) = 1$ for $\mu_{\text{adv}}$ regardless of the context. It is likely that $\mu_{\text{adv}}$ is not correlated with $\mu^*$ in measuring the true utility of the output.

Now, let us say we include two other versions of $y'$ obtained by applying the insertion and swapping perturbation function. In this case, $len(y) \leq len(y')$ will always be true; therefore, the value of $\mu_{adv}(x,y)$ will always be less than $\mu_{adv}(x,y')$. Hence, the adversarial metric will always achieve $ACC_\mu(Q_x) = 0$. If we aggregate the values from the different perturbations, we will get a $ACC_\mu(Q_x) = 0.33$.

\begin{figure*}
    \centering
    \begin{subfigure}[t]{0.33\textwidth}
        \centering    
        \includegraphics[width=\textwidth]{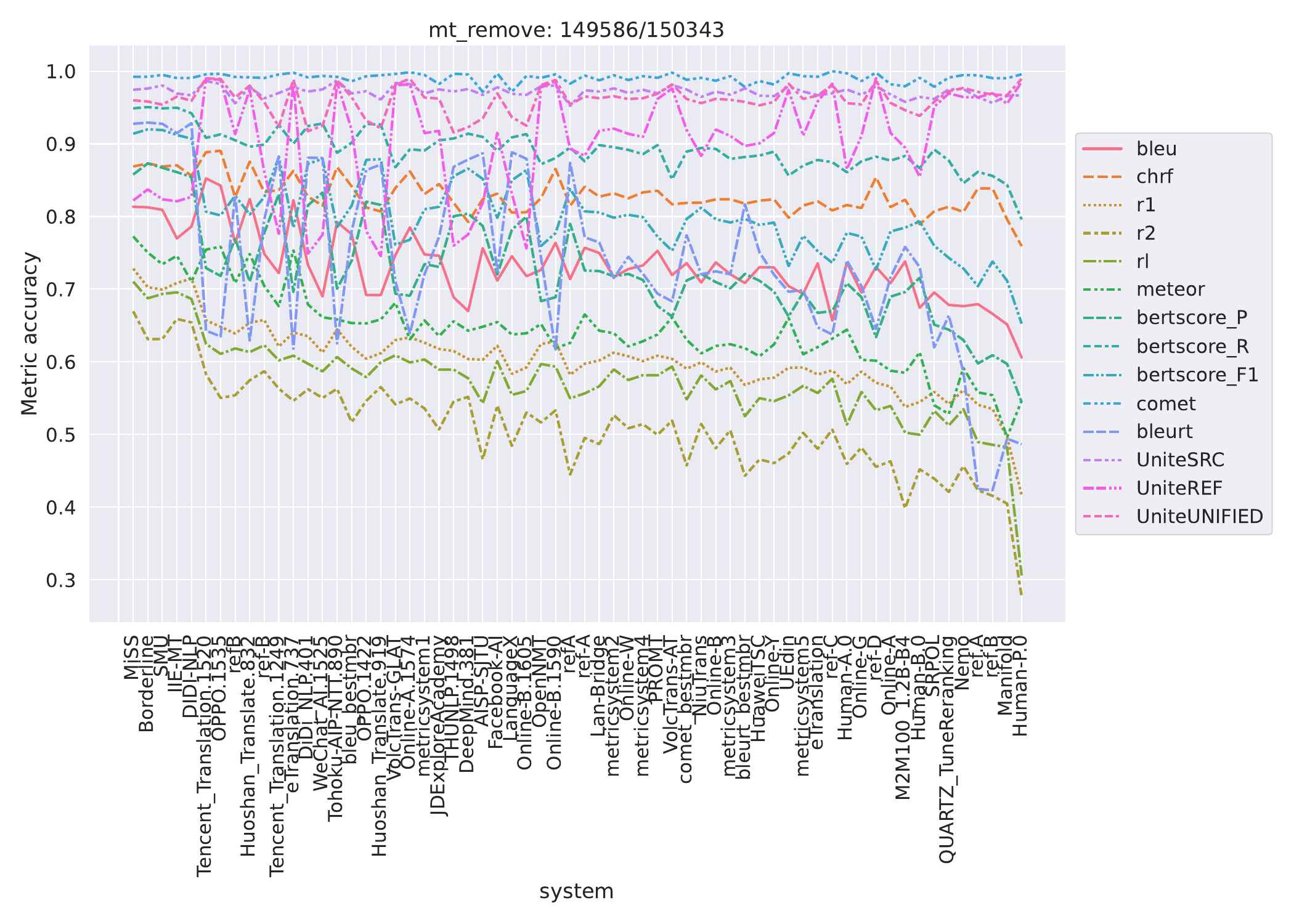}
        \caption{Removal}
    \end{subfigure}\hfill
    \begin{subfigure}[t]{0.33\textwidth}
        \centering    
        \includegraphics[width=\textwidth]{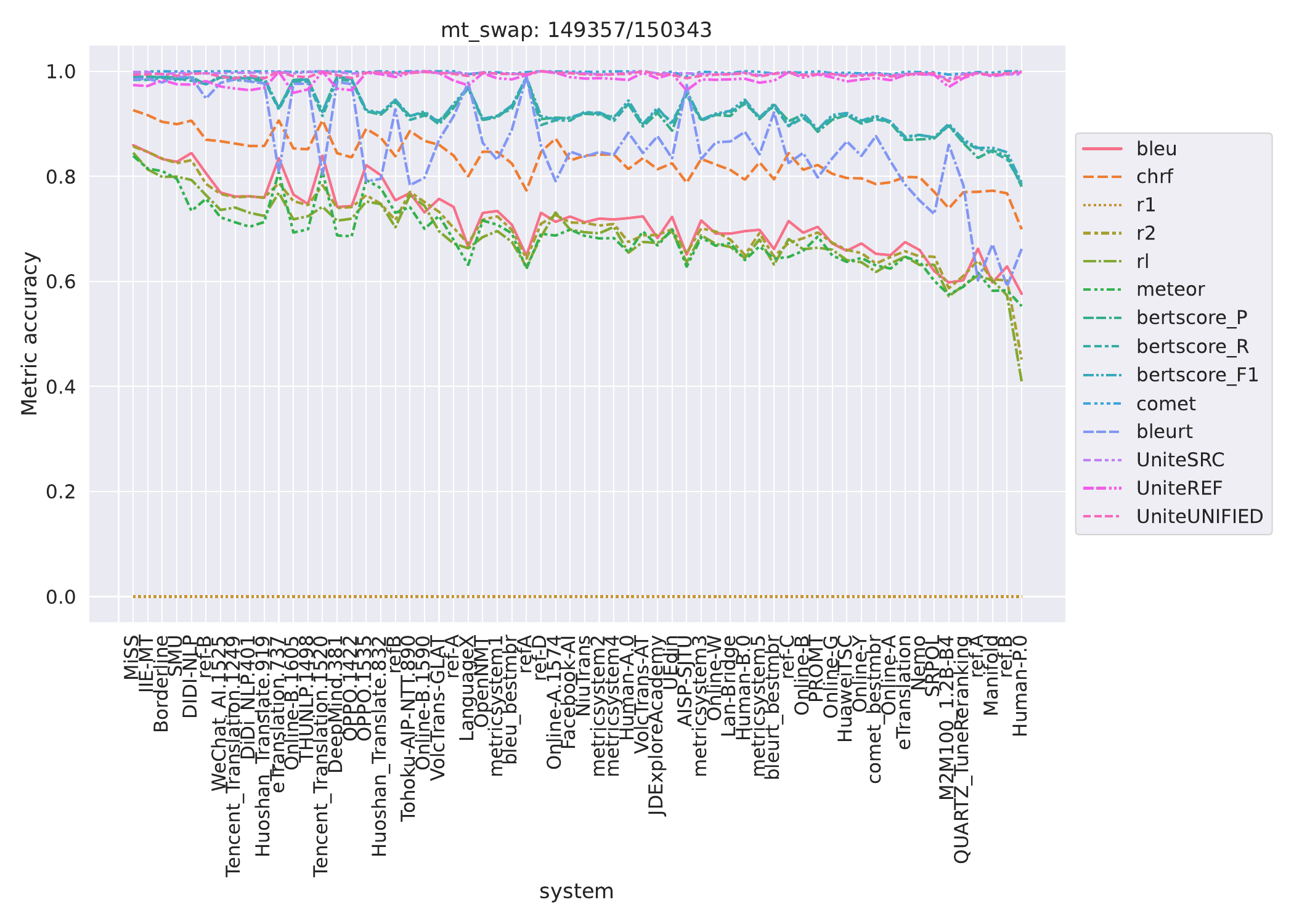}
        \caption{Swapping}
        \label{fig:mt_swap}
    \end{subfigure}\hfill
    \begin{subfigure}[t]{0.33\textwidth}
        \centering    
        \includegraphics[width=\textwidth]{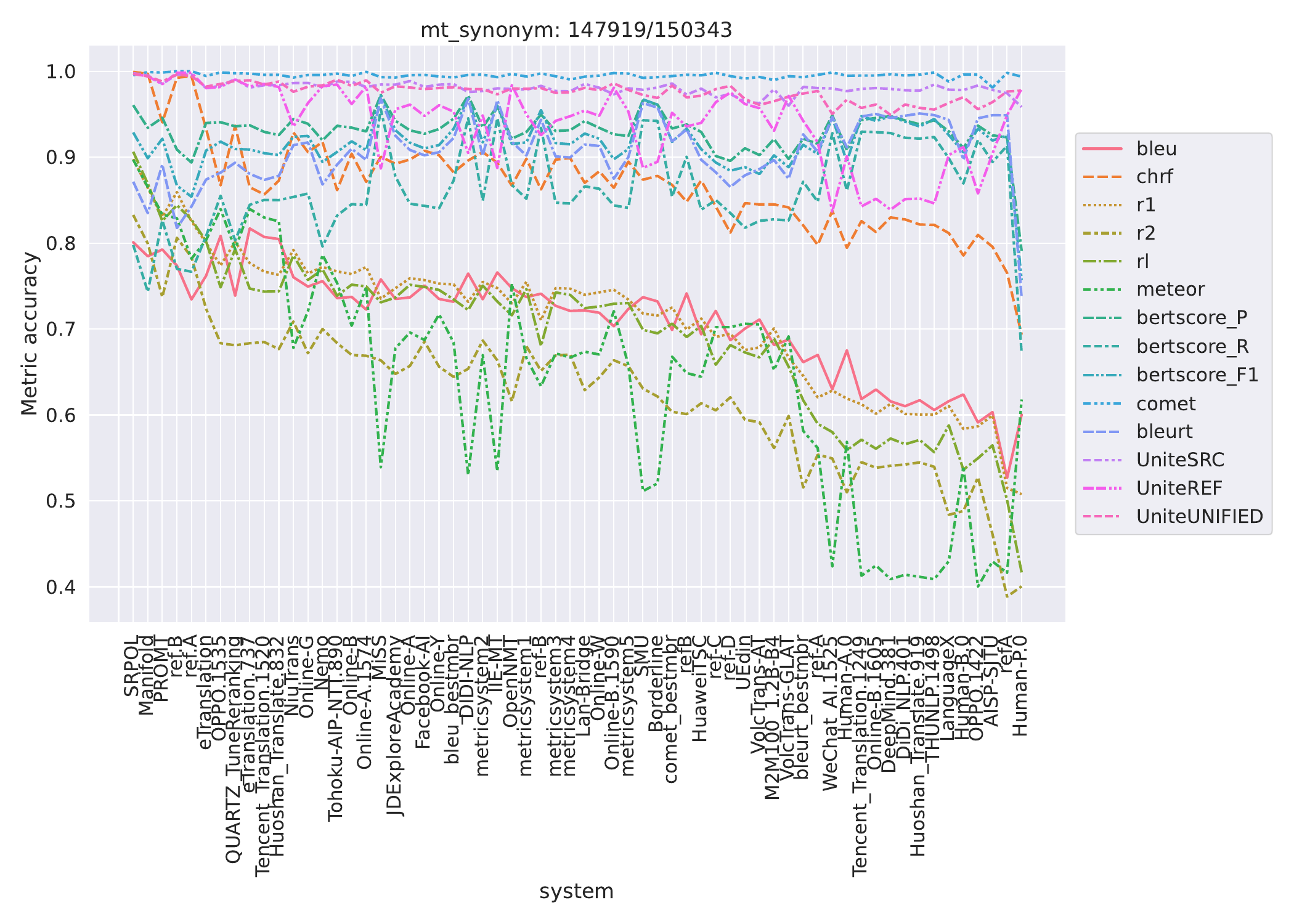}
        \caption{Synonym substitution}
    \end{subfigure}
    
    \vspace{0.3cm} 

    \begin{subfigure}[t]{0.33\textwidth}
        \centering    
        \includegraphics[width=\textwidth]{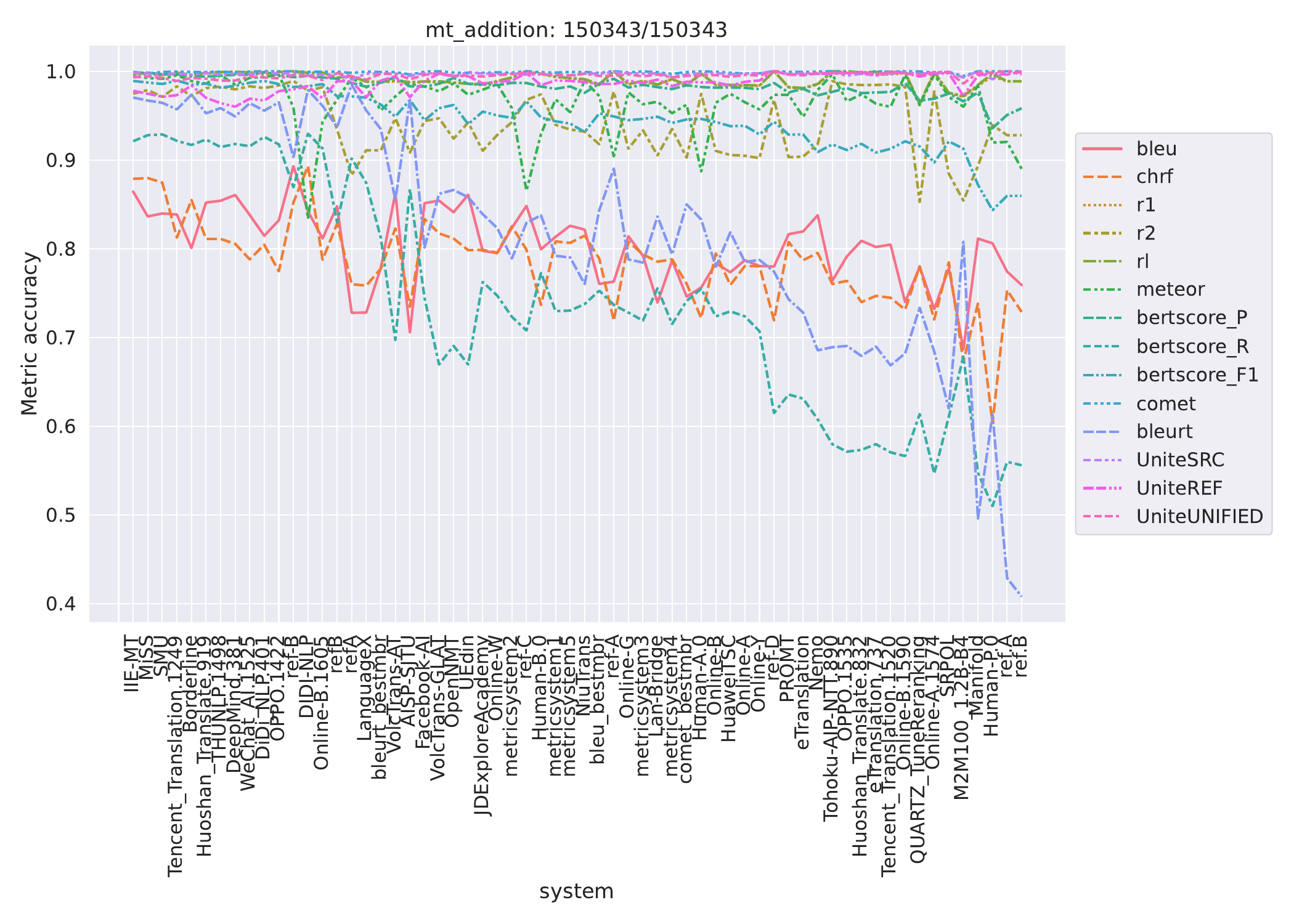}
        \caption{Insertion}
    \end{subfigure}\hfill
    \begin{subfigure}[t]{0.33\textwidth}
        \centering    
        \includegraphics[width=\textwidth]{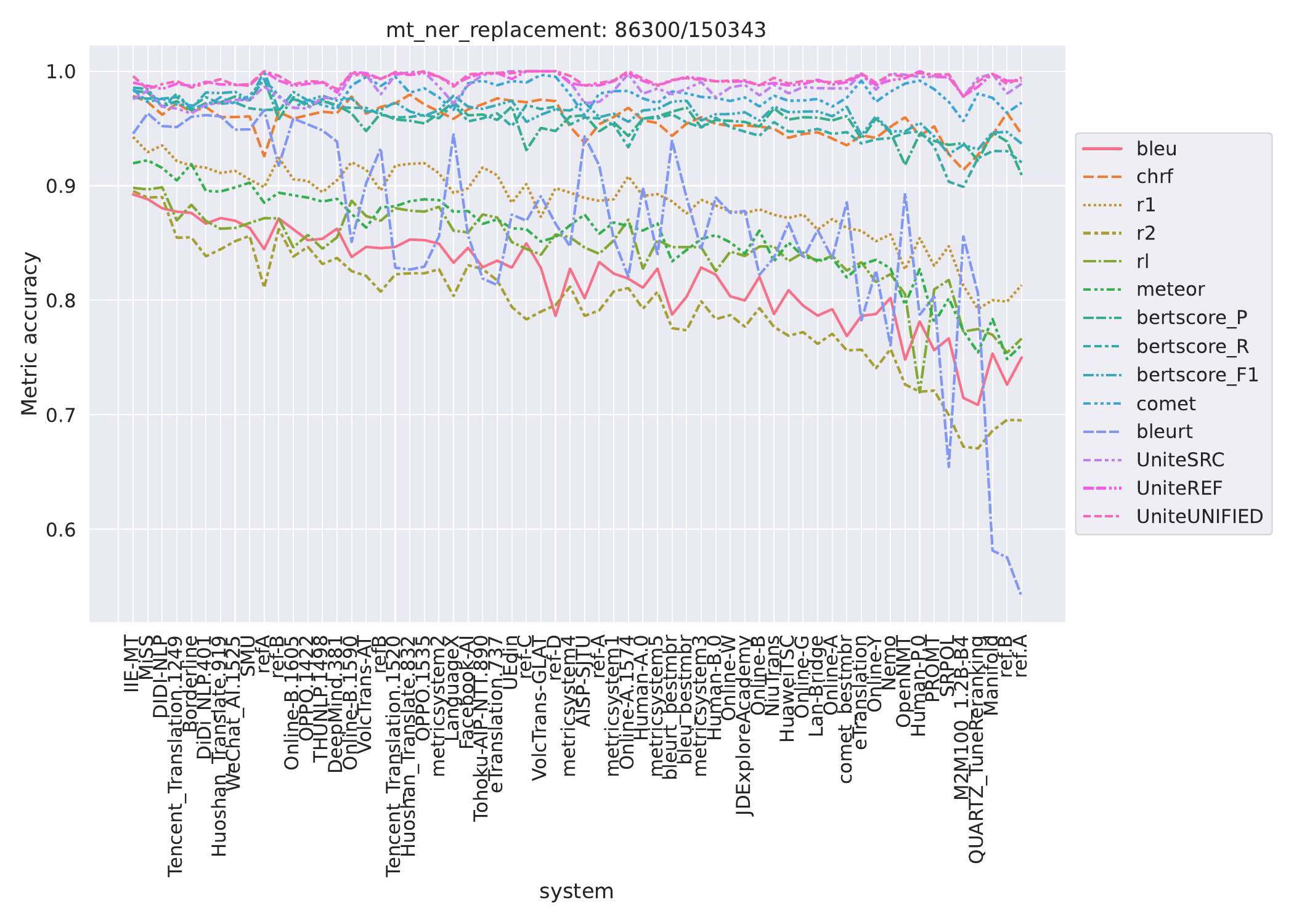}
        \caption{Named entity substitution}
    \end{subfigure}\hfill
    \begin{subfigure}[t]{0.33\textwidth}
        \centering    
        \includegraphics[width=\textwidth]{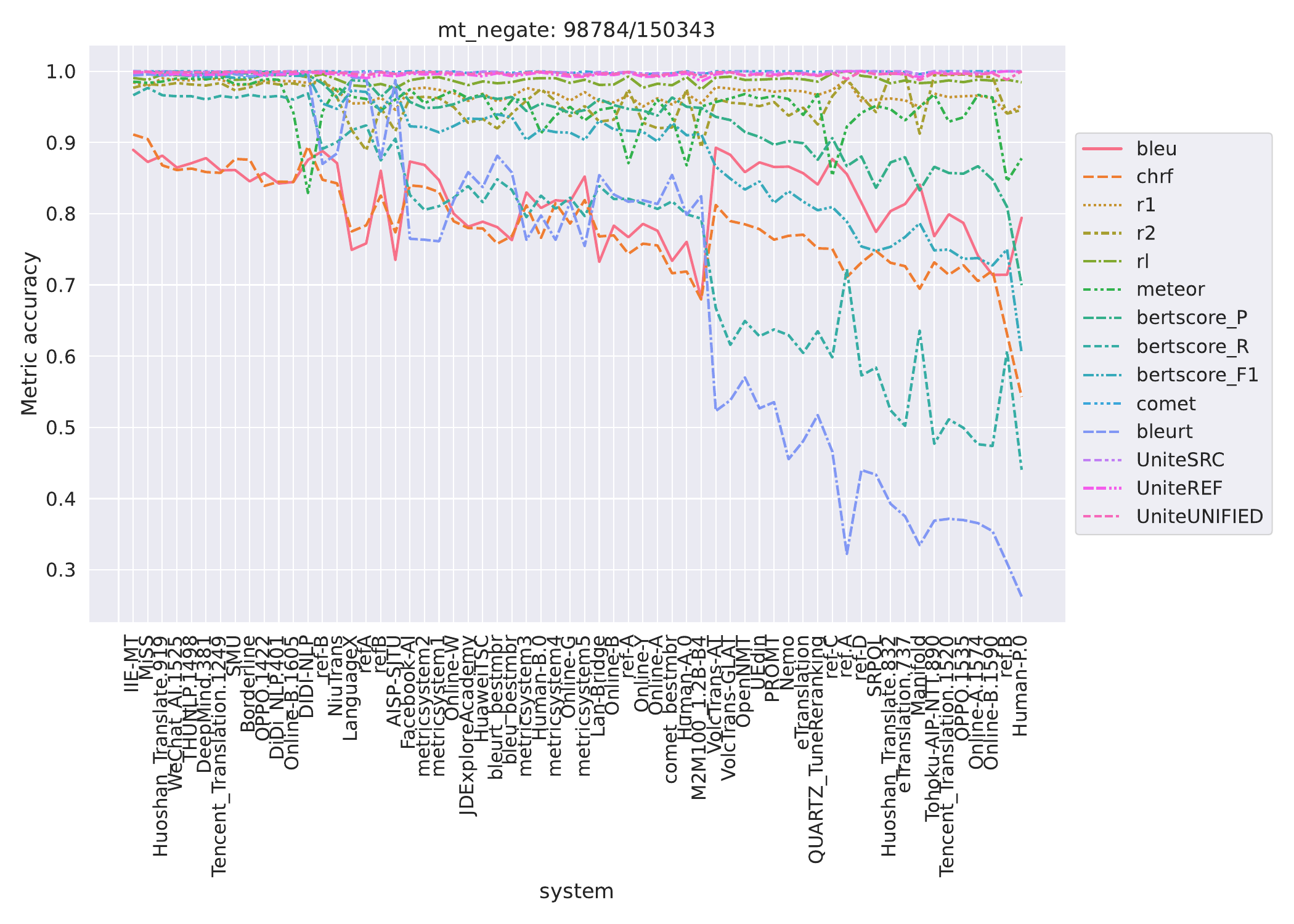}
        \caption{Negation}
    \end{subfigure}

    \vspace{0.3cm} 

    \begin{subfigure}[t]{0.33\textwidth}
        \centering    
        \includegraphics[width=\textwidth]{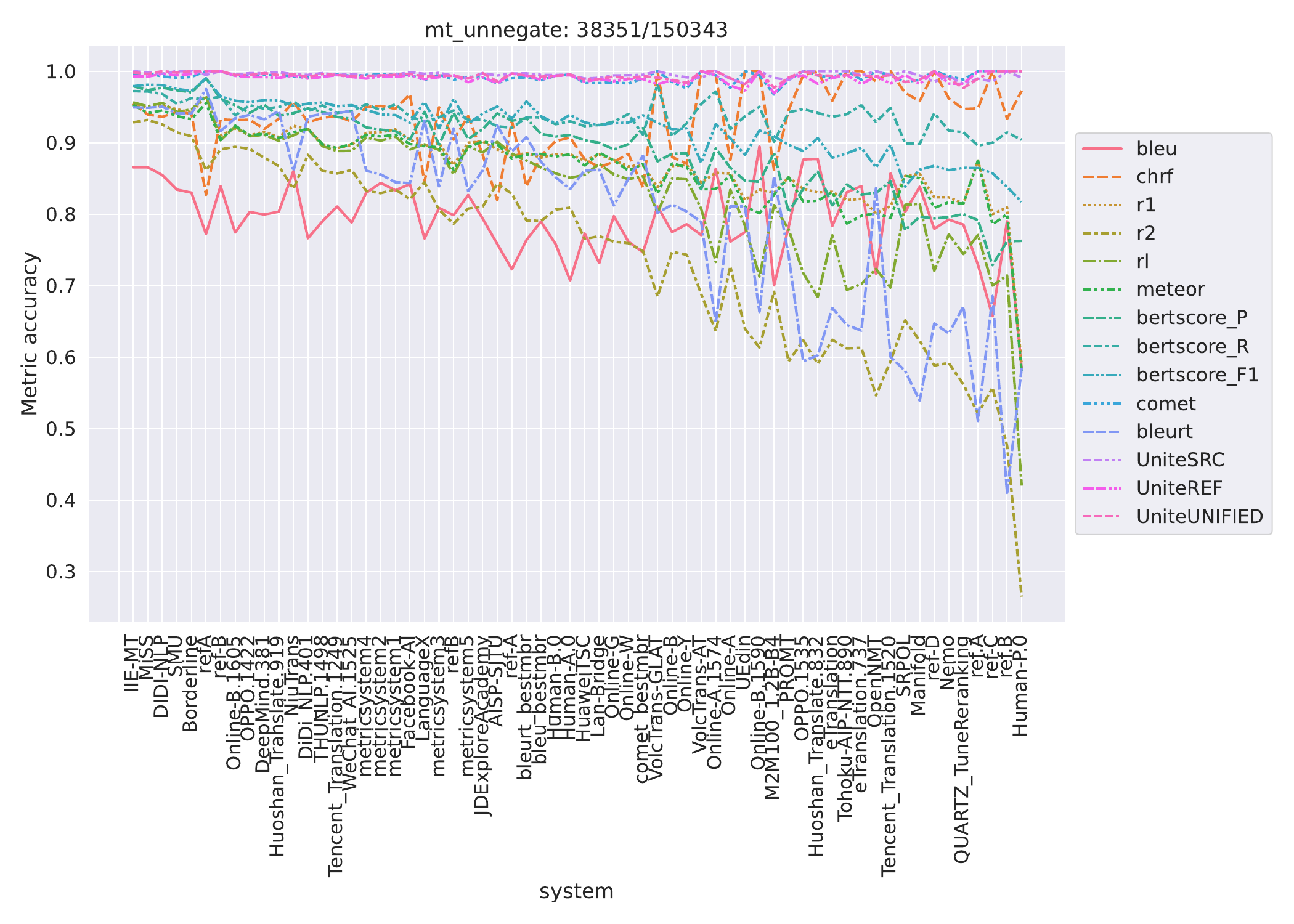}
        \caption{Negation removal}
    \end{subfigure}\hfill
    \begin{subfigure}[t]{0.33\textwidth}
        \centering    
        \includegraphics[width=\textwidth]{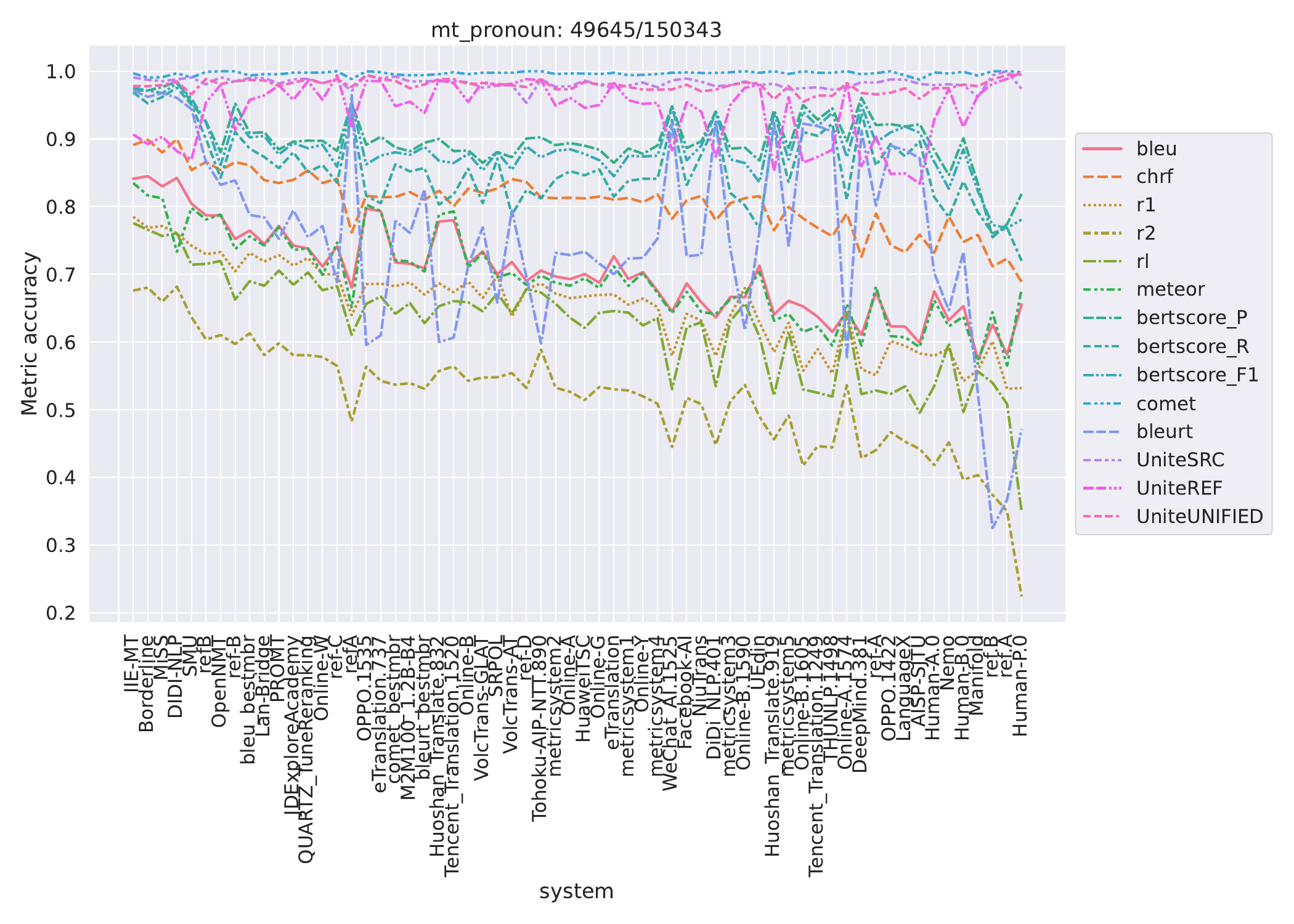}
        \caption{Pronoun substitution}
    \end{subfigure}\hfill
    \begin{subfigure}[t]{0.33\textwidth}
        \centering    
        \includegraphics[width=\textwidth]{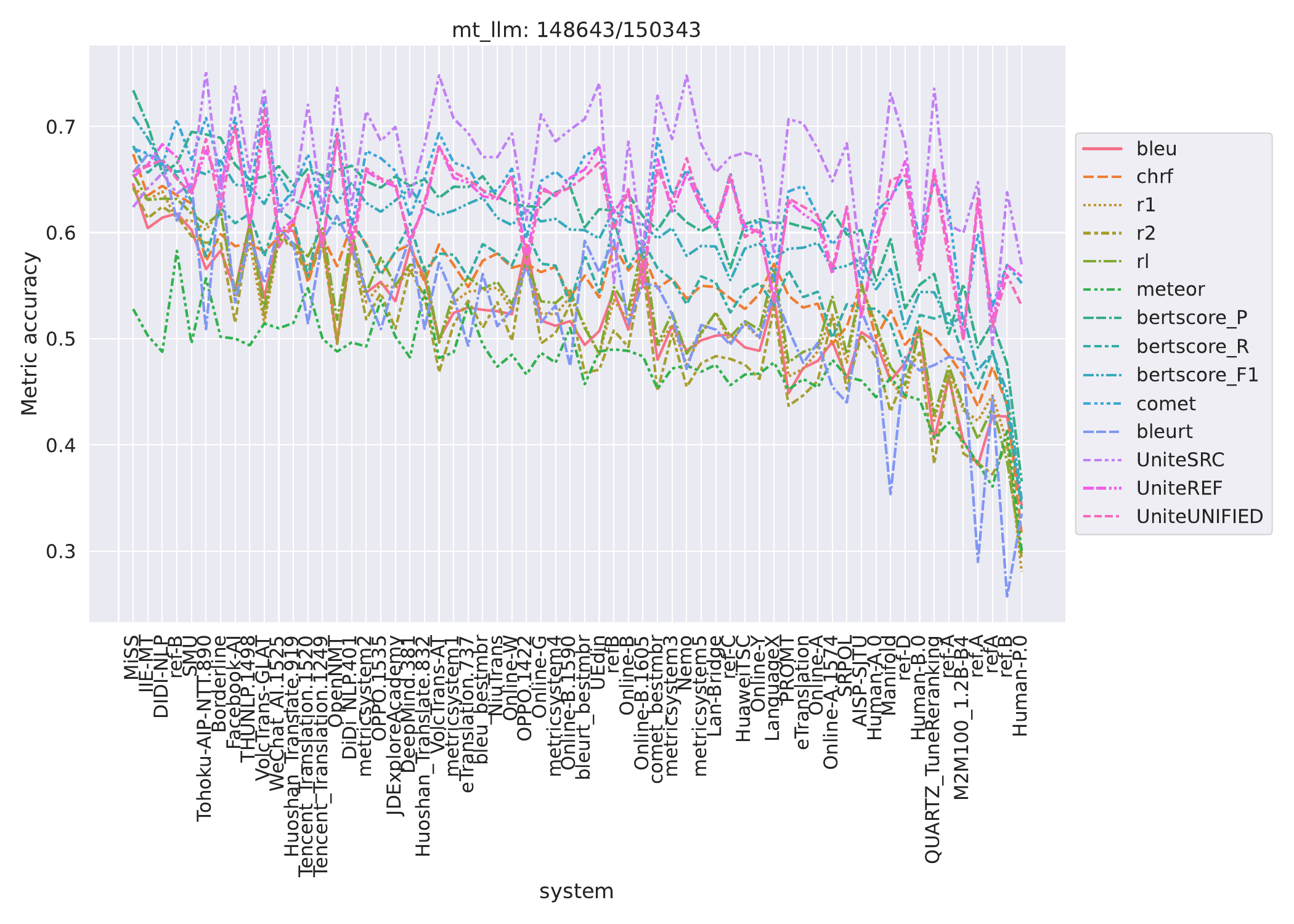}
        \caption{LLM-based perturbation}
    \end{subfigure}

    \caption{Metric accuracy graphs from using a single perturbation technique on each instance.}
    \label{fig:mt_perturbations}
\end{figure*}

We can observe this phenomenon from empirical examples of perturbing machine translation outputs using nine different techniques in Figure \ref{fig:mt_perturbations}. As discussed in Section \ref{sec:perturbation_techniques}, the swapping perturbation function penalizes disproportionately \textsc{Rouge-1}, as the metric measures the position-agnostic lexical alignment between two outputs. Consequently, we can see that in Figure \ref{fig:mt_swap}, the metric accuracy values for \textsc{Rouge-1} are consistently 0. However, when we observe its metric accuracy values under the other perturbation functions, they fluctuate at values above 0. If we compare this with other metrics, such as the family of \textsc{Unite} metrics, their metric accuracy values have a smaller variance between the different perturbation functions.

From the examples above, we can observe that if a metric confounds with specific perturbation methods, its local metric accuracy scores across different contexts would have a high variance when we change the perturbation method. This follows from the definition of variance: when a variable (in our case, the local metric accuracy under different perturbations) has a high variance, it means that its values are spread out over a wide range, indicating that under some perturbations the local metric accuracy is high, while it is low in others, instead of being close to the average value. Given that we keep the contexts the same when measuring the variances, we can also say that if the local metric accuracy of a particular metric across different contexts shows a high variance when changing the perturbation method, then it is likely caused by the presence of confounding between the perturbation and the metric design.

We will show empirically that this variance will be reduced if we combine different perturbation methods for machine translation. First, we obtain all the possible combinations of perturbations: each $k = [1,2,3,4,5,6,7,8]$ denoting the number of perturbations in each combination would have $9\choose k$ instances each. We omit $k=9$ because we would just have one combination of 9 different perturbations. For example, a combination of one perturbation method would be $9 \choose 1$$=9$ different single perturbations, while a combination of two perturbation methods would be $9 \choose 2$$=36$ different pairs, where each $y$ would have two different $y'$s, and so on. Let us denote the different groups of combinations $c_k$, where each $c_k$ contains all the possible perturbation combinations of size $k$. We randomly chose $\min($$9\choose k$$, 10)$ for each $c_k$ to speed up the calculation. For each combination in $c_{(k,i)}$, we compute the means of the local metric accuracies across the different machine translation systems. For each metric $\mu$, we compute how it varies with the local metric accuracies of the other possible perturbation combinations of size $k$. 

\begin{figure}[h]
 \includegraphics[width=0.49\textwidth]{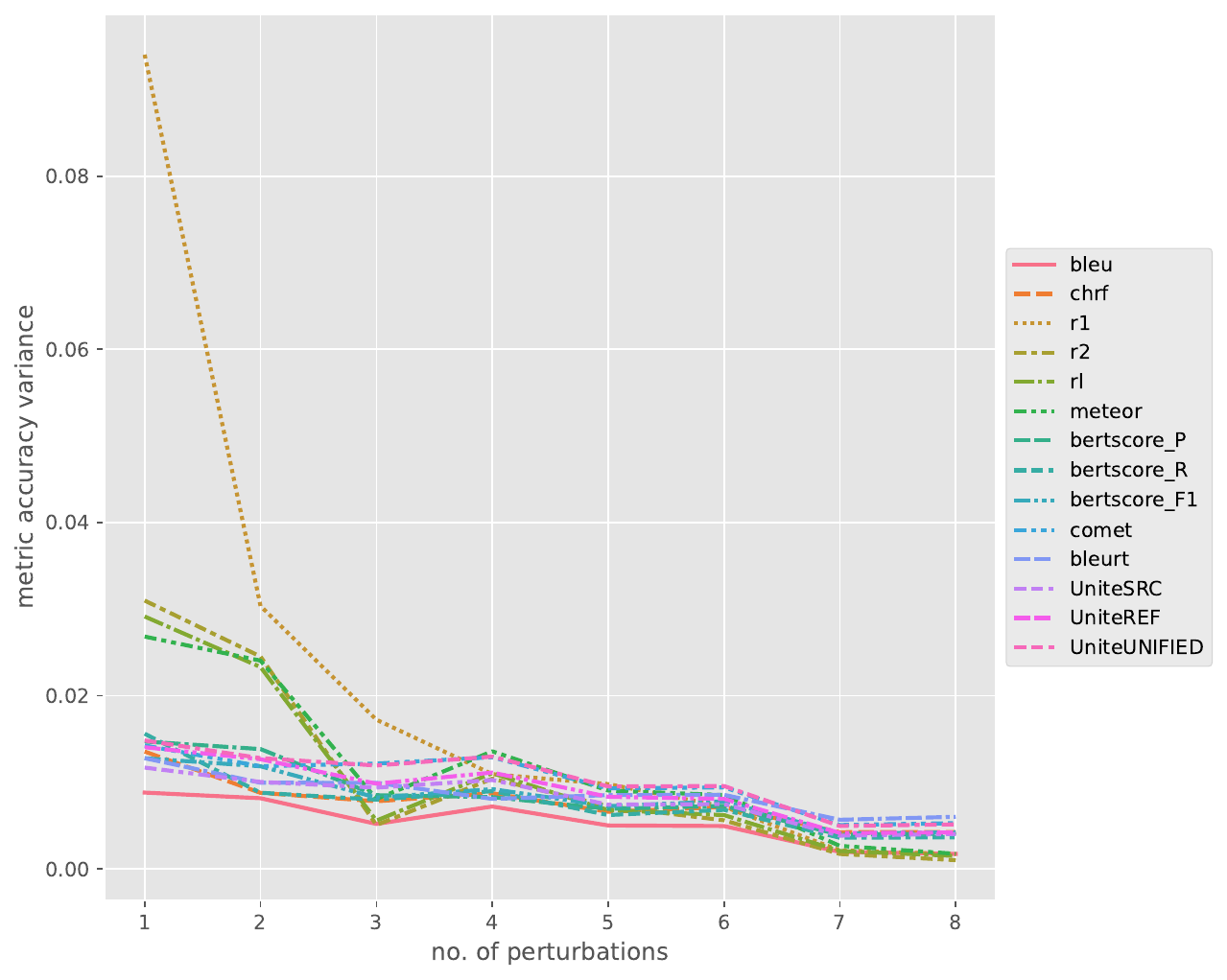}
 \caption{Local metric accuracy variance between the different number of perturbation combinations. }
 \label{fig:var_plot}
\end{figure}

We show the above result in Figure \ref{fig:var_plot}. We can interpret it as follows: the variance between groups of two perturbation combinations (for example, between $\tup{f_\text{removal}, f_\text{swap}}$ and $\tup{f_\text{removal}, f_\text{insertion}}$) is higher than between groups of three perturbation combinations (for example, between $\tup{f_\text{removal}, f_\text{swap}, f_\text{insertion}}$ and $\tup{f_\text{removal}, f_\text{LLM}, f_\text{negation}}$) for metrics such as \textsc{BleuRT} and \textsc{Rouge-1}. Therefore, we can see that by increasing the number of perturbations for a single instance of $y$, we will see reduced effects on confounders.

\subsection{Metric accuracy graphs}
\label{sec:other_categories}

\subsubsection{Machine translation}

\begin{figure}[h]
 \includegraphics[width=0.425\textwidth]{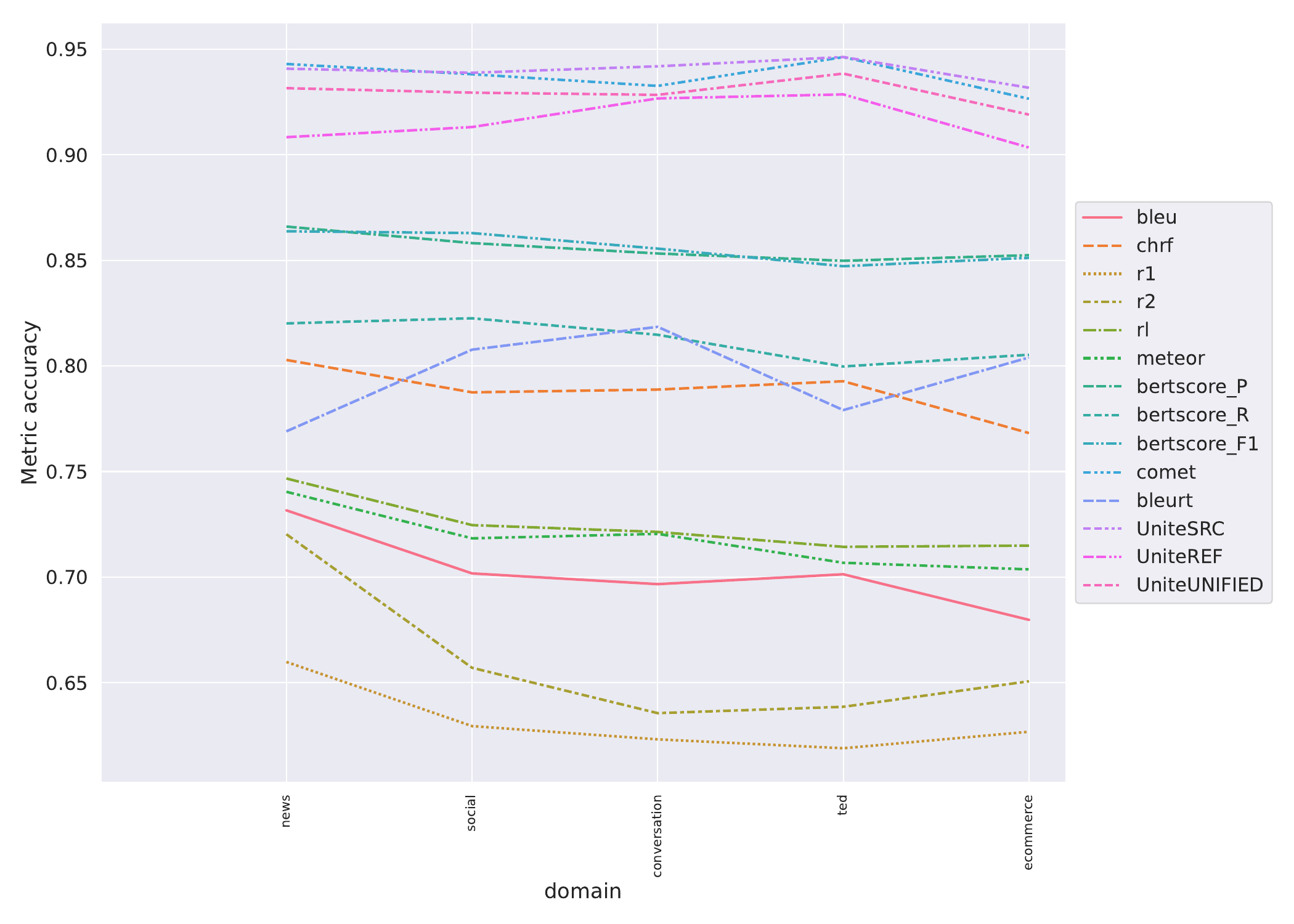}
 \caption{Metric accuracy for machine translation metrics across the different \textsc{Domain}s.}
 \label{fig:asr_systems}
\end{figure}

\begin{figure}[h]
 \includegraphics[width=0.425\textwidth]{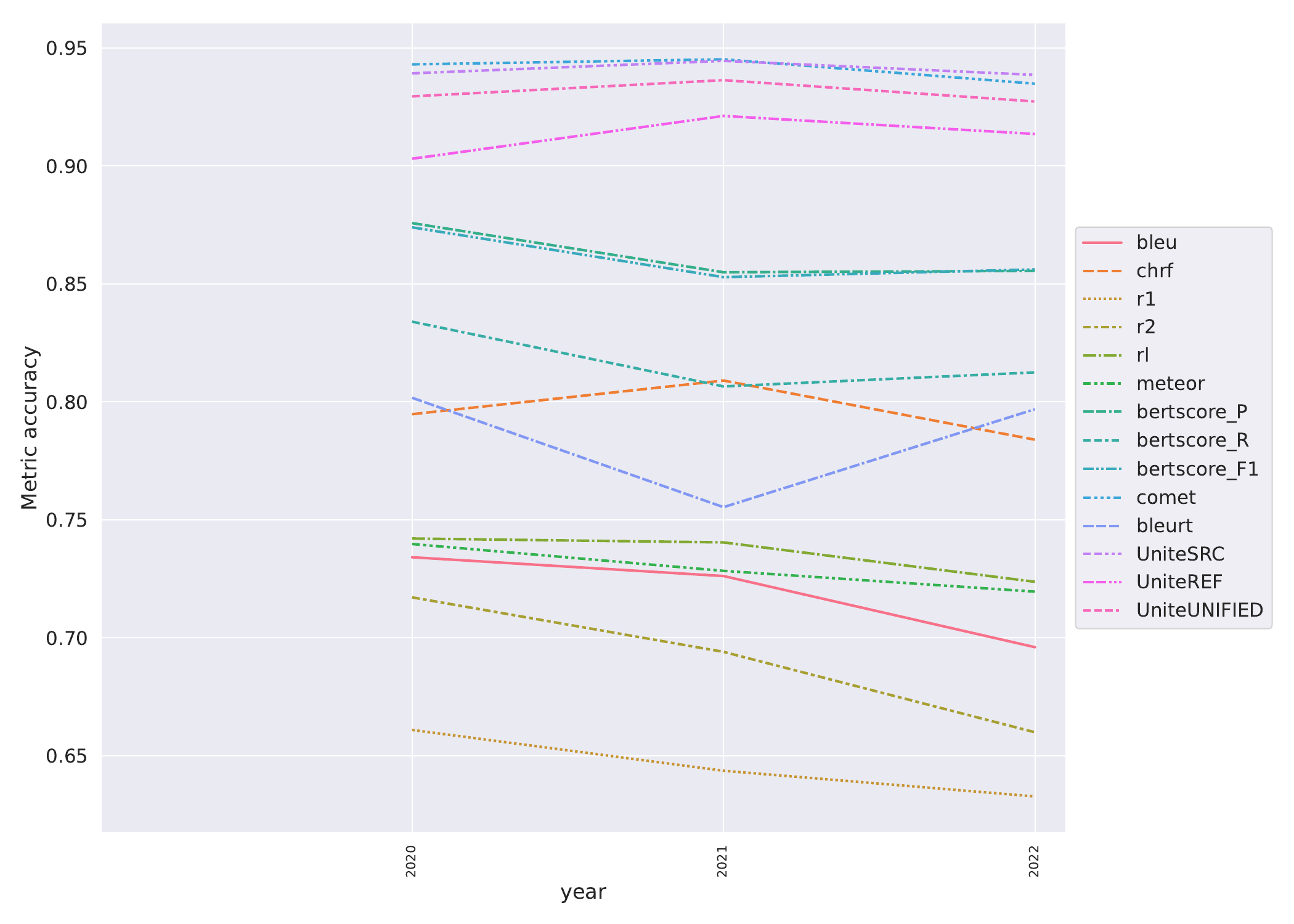}
 \caption{Metric accuracy for machine translation metrics across the different \textsc{Year}s.}
 \label{fig:asr_systems}
\end{figure}

\subsubsection{Automatic speech recognition}

\begin{figure}[h]
 \includegraphics[width=0.425\textwidth]{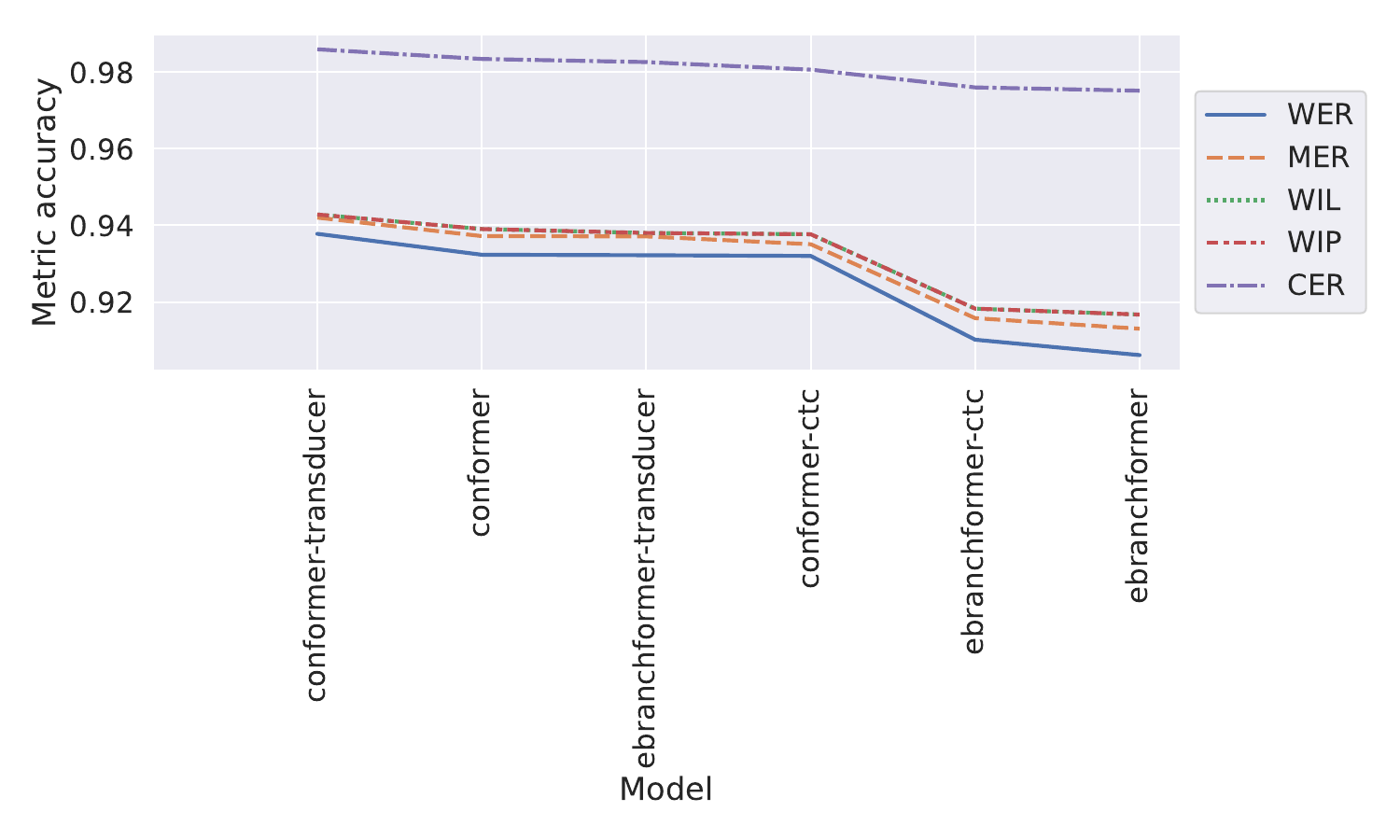}
 \caption{Metric accuracy for automatic speech recognition metrics across the different \textsc{System}s.}
 \label{fig:asr_systems}
\end{figure}

\subsection{Statistical Significance Testing results}
\label{sec:statsig_results}
The statistical significance test results referred to for machine translation (Section \ref{sec:mt_results}), automatic speech recognition (Section \ref{sec:asr_results}), and ranking (Section \ref{sec:ranking_results}), are shown in Tables \ref{tab:mt_statsig}, \ref{tab:asr_statsig}, \ref{tab:ranking_statsig}, respectively.

\begin{table}[h!]
\small
\centering
\begin{tabular}{lccc}
\textbf{Metric} & \textbf{$\chi^2$ stat} & \textbf{p-value} & \textbf{D.O.F.} \\
\hline
bleu    & 3528.549 & 0.0   & 61 \\
chrf    & 2821.829 & 0.0   & 61 \\
r1    & 1918.071 & 0.0   & 61 \\
r2    & 5709.353 & 0.0   & 61 \\
rl    & 3475.848 & 0.0   & 61 \\
meteor   & 2328.442 & 0.0   & 61 \\
bertscore\_P  & 4587.739 & 0.0   & 61 \\
bertscore\_R  & 9192.163 & 0.0   & 61 \\
bertscore\_F1  & 5485.387 & 0.0   & 61 \\
comet    & 702.717 & 7.264963e-110 & 61 \\
bleurt   & 19874.815 & 0.0   & 61 \\
UniteSRC   & 699.907 & 2.631570e-109 & 61 \\
UniteREF   & 2527.075 & 0.0   & 61 \\
UniteUNIFIED  & 834.976 & 2.211392e-136 & 61 \\
\end{tabular}
\caption{$\chi^2$ Contingency Test Results for machine translation metrics}
\label{tab:mt_statsig}
\end{table}

\begin{table}[h!]
\small
\centering
\begin{tabular}{lccc}

\textbf{Metric} & \textbf{$\chi^2$ stat} & \textbf{p-value} & \textbf{D.O.F.} \\
\hline
WER    & 677.508 & 2.884115e-99 & 72 \\
MER    & 561.242 & 7.168449e-77 & 72 \\
WIL    & 1328.151 & 2.412446e-230 & 72 \\
WIP    & 3956.386 & 0.0   & 72 \\
CER    & 1085.986 & 8.197217e-181 & 72 \\
\end{tabular}
\caption{$\chi^2$ Contingency Test Results for automatic speech recognition Metrics}
\label{tab:asr_statsig}
\end{table}

\begin{table}[h!]
\small
\centering
\begin{tabular}{lccc}
\textbf{Metric} & \textbf{$\chi^2$ stat} & \textbf{p-value} & \textbf{D.O.F.} \\
\hline
map     & 154124.458 & 0.0 & 20 \\
Rprec     & 105742.720 & 0.0 & 20 \\
recip\_rank   & 47484.273 & 0.0 & 20 \\
iprec\_at\_recall\_0.00 & 57663.527 & 0.0 & 20 \\
iprec\_at\_recall\_0.10 & 178593.017 & 0.0 & 20 \\
iprec\_at\_recall\_0.20 & 237843.573 & 0.0 & 20 \\
iprec\_at\_recall\_0.30 & 234700.982 & 0.0 & 20 \\
iprec\_at\_recall\_0.40 & 179188.961 & 0.0 & 20 \\
P\_5     & 122190.649 & 0.0 & 20 \\
P\_10     & 143434.121 & 0.0 & 20 \\
P\_15     & 147076.902 & 0.0 & 20 \\
P\_20     & 146921.819 & 0.0 & 20 \\
P\_30     & 140459.426 & 0.0 & 20 \\
recall\_5    & 122190.649 & 0.0 & 20 \\
recall\_10   & 143434.121 & 0.0 & 20 \\
recall\_15   & 147076.902 & 0.0 & 20 \\
recall\_20   & 146921.819 & 0.0 & 20 \\
recall\_30   & 140459.426 & 0.0 & 20 \\
ndcg\_cut\_5   & 148042.996 & 0.0 & 20 \\
ndcg\_cut\_10   & 173763.267 & 0.0 & 20 \\
ndcg\_cut\_15   & 179134.760 & 0.0 & 20 \\
ndcg\_cut\_20   & 179938.561 & 0.0 & 20 \\
ndcg\_cut\_30   & 175786.549 & 0.0 & 20 \\

\end{tabular}
\caption{$\chi^2$ Contingency Test Results for ranking metrics}
\label{tab:ranking_statsig}
\end{table}

\clearpage
\subsection{Metric accuracies for different translation qualities}

    For the machine translation metrics, we can divide the data into different translation quality contexts as the WMT 2022 dataset \cite{freitag:wmt2022:wmt} comes with MQM scores \cite{freitag:mqm:tacl}, which reflect the human judgment of the system outputs. The MQM scores for the English to German (\textsc{En-De}) and Chinese to English (\textsc{Zh-En}) translation pairs were mostly annotated by Google and ranges from -25 to 0, where 0 is a perfect translation, and -25 is the worst possible score. On the other hand, the MQM scores for the English to Russian (\textsc{En-Ru}) translation pairs were annotated by Unbabel and ranges from -inf to 100, where 100 is a perfect translation and something below 0 is a bad translation. We have used the data made available on Huggingface\footnote{\url{https://huggingface.co/datasets/RicardoRei/wmt-mqm-human-evaluation}}. 
    
    Because MQM scores are continuous, we discretized the scores into 250 buckets of equal frequency. We did this instead of discretizing the scores into equal ranges, as some ranges of MQM scores had no data points in them.

\begin{figure}[h]
 \includegraphics[width=0.48\textwidth]{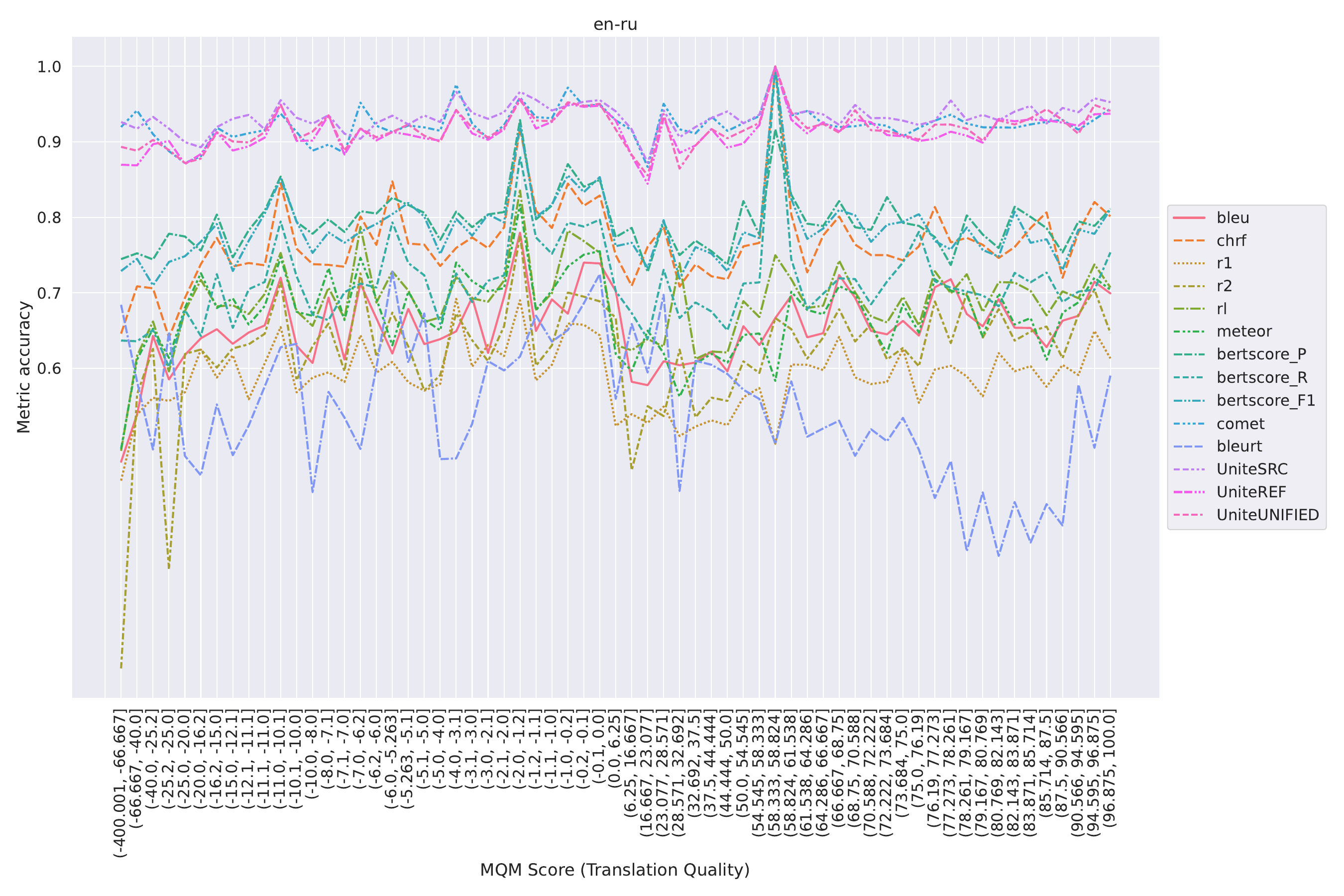}
 \caption{Local metric accuracy across the different MQM scores for English to Russian (\textsc{En-Ru}) translation pairs}
 \label{fig:mqm_enru}
\end{figure}

\begin{figure}[h]
 \includegraphics[width=0.48\textwidth]{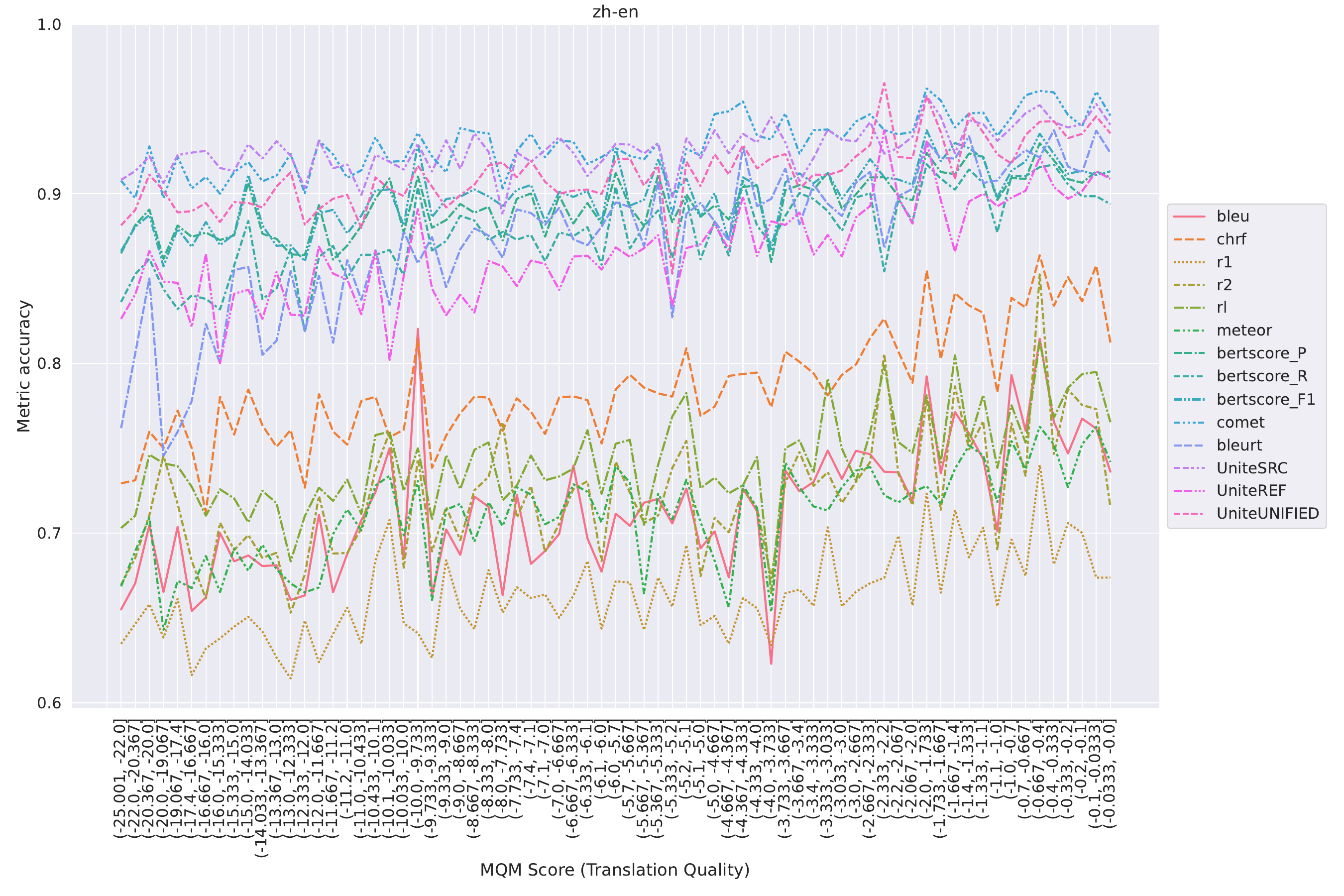}
 \caption{Local metric accuracy across the different MQM scores for Chinese to English (\textsc{Zh-En}) translation pairs}
 \label{fig:mqm_zhen}
\end{figure}

    In Section \ref{sec:mt_results}, we show the metric accuracies for English to German (\textsc{En-De}) translation pairs. We show similar results for English to Russian (\textsc{En-Ru}) and Chinese to English (\textsc{Zh-En}) translation pairs in Figures \ref{fig:mqm_enru} and \ref{fig:mqm_zhen}, respectively.

\newpage
\label{sec:appendix_mqmfigs}

\end{document}